\documentclass[twocolumn,journal]{IEEEtran}
\usepackage[T1]{fontenc}
\usepackage[utf8]{inputenc}
\usepackage{array}
\usepackage{float}
\usepackage{textcomp}
\usepackage{amsmath}
\usepackage{graphicx}
\usepackage{adjustbox}
\usepackage{subscript}
\usepackage{multirow}
\usepackage{nicefrac}
\usepackage[unicode=true,
 bookmarks=true,bookmarksnumbered=true,bookmarksopen=true,bookmarksopenlevel=1,
 breaklinks=false,pdfborder={0 0 0},pdfborderstyle={},backref=false,colorlinks=false]
 {hyperref}
\hypersetup{pdftitle={Your Title},
 pdfauthor={Your Name},
 pdfpagelayout=OneColumn, pdfnewwindow=true, pdfstartview=XYZ, plainpages=false}
\usepackage{breakurl}
\usepackage[colorinlistoftodos]{todonotes}
\usepackage{url}

\makeatletter

\providecommand{\tabularnewline}{\\}

\let\oldforeign@language\foreign@language
\DeclareRobustCommand{\foreign@language}[1]{%
  \lowercase{\oldforeign@language{#1}}}

\usepackage[caption=false,font=footnotesize]{subfig}
\usepackage{pgfplots}

\@ifundefined{showcaptionsetup}{}{%
 \PassOptionsToPackage{caption=false}{subfig}}
\usepackage{subfig}
\makeatother

\global\long\def\SystemName{FVV Live}

\begin{document}

\title{FVV Live: A real-time free-viewpoint video system with consumer electronics hardware}

\author{Pablo Carballeira, Carlos Carmona, C\'esar D\'iaz, Daniel Berj\'on, Daniel Corregidor, Juli\'an Cabrera, Francisco Mor\'an, Carmen Doblado, Sergio Arnaldo, Mª del Mar Mart\'in and Narciso Garc\'ia
\thanks{This work has been partially supported by the Ministerio de Ciencia, Innovaci\'on y Universidades (AEI/FEDER) of the Spanish Government under project TEC2016-75981 (IVME) and by Huawei Technologies Co., Ltd.

P. Carballeira was with the Grupo de Tratamiento de Im\'agenes, Universidad Polit\'ecnica de Madrid. He is now with the Video Processing and Understanding Lab, Escuela Polit\'ecnica Superior, Universidad Aut\'onoma de Madrid, Madrid 28049, Spain (e-mail: pablo.carballeira@uam.es)

C. Carmona, C. D\'iaz, D. Berj\'on, D. Corregidor, J. Cabrera, F. Mor\'an, C. Doblado, S. Arnaldo, M. Mart\'in and N. Garc\'ia are with the Grupo de Tratamiento de Im\'agenes, Information Processing and Telecommunications Center and ETSI Telecomunicaci\'on, Universidad Polit\'ecnica de Madrid, Madrid 28040, Spain (e-mail: \{ccv,cdm,dbd,dcl,julian.cabrera,fmb,cdo,sad,mmp,narciso\}@gti.ssr.upm.es).

This work has been submitted to the IEEE for possible publication. Copyright may be transferred without notice, after which this version may no longer be accessible.
}}


\maketitle

\begin{figure*}[ht!]
\centering{}\includegraphics[width=0.9\linewidth]{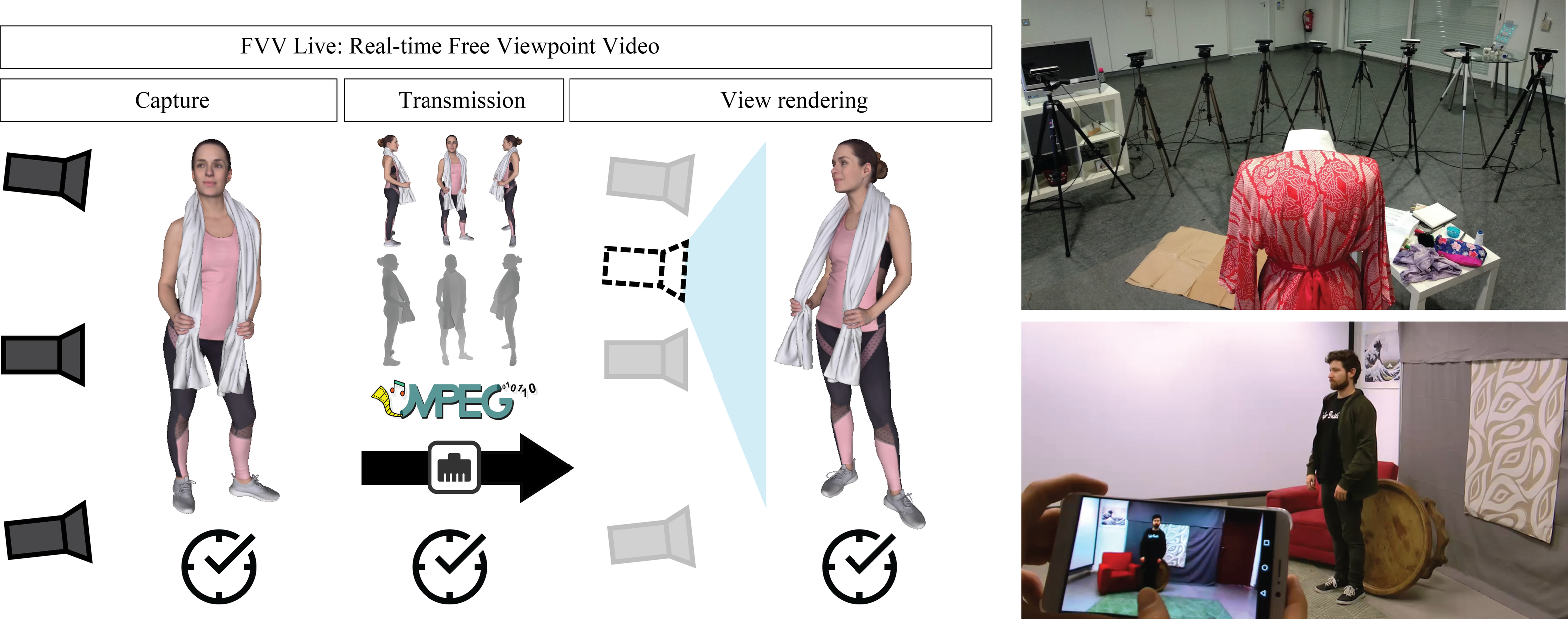}
\caption{\label{fig:FTV-Wei-complete-system} Concept of the \SystemName\ system: real-time free-viewpoint video (left). \SystemName\ camera setup (right-top). \SystemName\ rendering result displayed on the mobile screen (right-bottom).}
\end{figure*}

\begin{abstract}
\SystemName\ is a novel end-to-end free-viewpoint video system, designed for low cost and real-time operation, %
based on off-the-shelf components. The system has been designed to yield high-quality free-viewpoint video using consumer-grade cameras and hardware, which enables low deployment costs and easy installation for immersive event-broadcasting or videoconferencing.

The paper describes the architecture of the system, including acquisition and encoding of multiview plus depth data in several capture servers and virtual view synthesis on an edge server. All the blocks of the system have been designed to overcome the limitations imposed by hardware and network, which impact directly on the accuracy of depth data and thus on the quality of virtual view synthesis. The design of \SystemName\ allows for an arbitrary number of cameras and capture servers, and the results presented in this paper correspond to an implementation with nine stereo-based depth cameras.  

\SystemName\ presents low motion-to-photon and end-to-end delays, which enables seamless free-viewpoint navigation and bilateral immersive communications. Moreover, the  visual quality of \SystemName\ has been assessed through subjective assessment with satisfactory results, and additional comparative tests show that it is preferred over state-of-the-art DIBR alternatives.   
\end{abstract}


\IEEEpeerreviewmaketitle{}

\section{Introduction}

\IEEEPARstart{I}{mmersive} video technologies have experienced a considerable development over the last decade. One of these technologies is free-viewpoint video (FVV), a.k.a. free-viewpoint television, which allows the user to freely move around the scene, navigating along an arbitrary trajectory as if there were a virtual camera that could be positioned anywhere within the scene. This functionality can improve the user experience in event broadcasting, such as sports~\cite{suenaga2015practical}~\cite{sabirin2018semi} or  performances (theater, circus, etc.), and in interactive video communications such as immersive videoconferences~\cite{microsoft_holoportation} or interactive courses (medicine, dance, etc.).

Typically, any FVV system requires three main blocks~\cite{SMOLIC20111958} (see Fig.~\ref{fig:FTV-Wei-complete-system}): (i) a volumetric video acquisition stage that captures the scene from multiple viewpoints and converts the captured data into a 3D representation of the scene containing both texture and geometric information; (ii) a compression and transmission module; and (iii) a view synthesis block that is in charge of rendering the view  according to the desired position of the virtual camera \cite{stankiewicz2018free}.

The development of such systems presents several challenges regarding video quality, real-time operation  and cost, which are often antagonistic. High-quality volumetric video benefits from the quality and number of cameras in the video acquisition stage, which in turn increases the deployment costs of the acquisition setup, and requires more resources in terms of processing capabilities (compression and rendering) and network bandwidth. Additionally, high-quality volumetric video typically requires synthesis algorithms of such a high computational complexity that prevent its real-time operation despite the use of high-end computation resources. 

Commercial systems such as Intel True View~\cite{intel_true_view} or 4DReplay \cite{4d_replay} have oriented their application towards the quality end, sacrificing real-time operation and user interactivity (the virtual navigation path is predefined at the acquisition side). Telepresence systems~\cite{mekuria17} typically focus on the reconstruction of human shapes, prioritizing real-time operation over video quality.

In this paper, we present a novel end-to-end real-time FVV system, \SystemName, that has been designed to provide high virtual video quality using off-the-shelf hardware, thus enabling low-cost and easy deployment. The key elements of this system are the following:      

\begin{itemize}
    \item An acquisition block comprised by a sparse array of consumer electronics stereo cameras, managed by a set of capture servers (CSs), which yields a multiview plus depth (MVD) format. The acquisition block includes a depth post-processing module that deals with depth estimation errors due to slight calibration errors typical in stereo-based depth cameras. 
    
    \item A compression and transmission block based on standard video coding schemes and transmission protocols. The design of the compression block focuses on preserving depth data to improve synthesis quality. To limit the overall bitrate, the transmission block adaptively enables/disables the data stream from each real camera depending on the position of the virtual one.
    
    \item A view synthesis module providing high quality video with real-time constraints. This module runs at a single edge server (ES) and uses a layered approach, merging background (BG) and foreground (FG) layers projected from several reference cameras in the virtual view.

\end{itemize} 
    
The layered design of the view synthesis module requires a real-time FG/BG segmentation performed in the CSs, and drives as well the design of other elements of the system. The motion degree asymmetry between BG and FG is exploited to save computing resources and bitrate by encoding and transmitting only time-varying depth data. These bitrate savings allow us to re-allocate bandwidth to transmit high-fidelity FG depth information, using 12-bit lossless coding schemes for the depth, which improves synthesis quality.

The design of \SystemName\ allows scalability for an arbitrary number of  cameras and CSs, and here we report results for an implementation with nine Stereolabs ZED cameras~\cite{zed} managed by three CSs. The MVD compressed streams of all cameras are sent to a single ES that renders the synthesized view corresponding to the virtual viewpoint selected by the user. 
The capture setting and real-time virtual view rendering are illustrated in Fig.~\ref{fig:FTV-Wei-complete-system}. In our implementation, the virtual viewpoint is selected continuously through a swiping interface in a smartphone, that displays the virtual view rendered at the ES. 

\SystemName\ is able to operate in real time at 1920x1080p resolution and 30~fps using off-the-self components (cameras and hardware), with and end-to-end delay of around 250~ms, and a mean motion-to-photon latency (i.e., time required to update the virtual view as a response to a viewpoint change) below 50~ms. These low delay values allow an immediate and natural navigation of the scene, and bilateral immersive communications. The visual quality of \SystemName\ has been assessed by means of a thorough subjective evaluation, which includes a comparison with state-of-the-art view synthesis algorithms. The results show that the visual quality of \SystemName\ is entirely satisfactory, and overwhelmingly preferred over other view synthesis alternatives for different camera densities in the camera setting and virtual view trajectories (see \texttt{\url{https://www.gti.ssr.upm.es/fvvlive}} for video demo).  \SystemName\ has been awarded with the 2019 Technology Award of the IET Vision and Imaging Network and accepted in 2020 ICME's demo track~\cite{fvv_live_icme}.

\section{Previous Work}

\noindent The most prominent use of FVV systems can be found in the field of sport replays. Examples of widely used commercial systems are 4DReplay~\cite{4d_replay} or Intel True View Technology~\cite{intel_true_view}. 4DReplay provides video replays for sports, events and film scenes in which the camera virtually travels along a predefined path. Their system uses more than 100 professional-grade cameras with an inter-camera distance between 5 and 10 cm and hardware synchronization. Intel True View technology uses an array of 30-50 high-end 5K cameras to capture volumetric data (voxels) of all the action. Those voxels allow the technology to render replays in multi-perspective 3D.  

Both systems can be framed in the high-quality end of FVV systems, requiring a dense array of professional cameras to produce high quality video replays. However, the FVV functionality is limited, due to the use of predefined virtual paths and non real-time operation. Moreover, virtual paths are typically designed to improve perceptual quality, rapidly swiping trough virtual views and stopping in viewpoints of physical cameras, which masks rendering artifacts. In the case of 4DReplay, the density of the camera arrays allows to generate the viewpoint path from consecutive frames from adjacent physical cameras, without requiring virtual view synthesis.  %

The development of a FVV system that works in real time, yielding high quality virtual video for arbitrary virtual paths, and with low deployment costs, still presents a great challenge that we have addressed in this work. Here, we review existing approaches, and discuss the advantages and limitations, of the three main blocks of any FVV system, namely acquisition, compression and transmission, and rendering.

\subsection{Acquisition}
\label{subsec:capture_sota}
\noindent The goal of the acquisition block in a FVV system is to yield a data format that includes multiview texture and geometry information, enabling the system to render the scene from an arbitrary viewpoint. FVV data formats are closely related to rendering techniques, and span from purely image-based formats such as lightfields, that implicitly encode the scene geometry, to explicit 3D mesh models~\cite{SMOLIC20111958} or point clouds~\cite{berger2017survey}. 

Typically, the geometry information is first estimated from multiple viewpoints, in the form of depth data. Then, this multiview geometry information can be transmitted independently, as in the MVD format, or combined into a single geometry format such as a 3D mesh~\cite{microsoft_fvv}. Intermediate formats exist, such as layered depth video~\cite{LDV}, in which the occluded areas of the primary depth view are supplemented with depth data from other views.

Different technologies allow to obtain depth data in real time, and can be classified into active vs. passive, depending on whether the devices they use emit signals onto the scene or not~\cite{active_passive_3D}. Time of flight (ToF) cameras \cite{kinect_v2}, LIDAR sensors~\cite{velodyne} and structured light cameras~\cite{realsense} are active depth sensing devices, while stereo~\cite{zed} and plenoptic~\cite{wu2017light} cameras are passive depth estimation devices.

Even if active devices differ in their spatial resolution, depth range, frame rate or operation conditions, their key advantage is generally a good accuracy in the depth measurements. On the other hand, their active nature typically prevents a multi-camera setting due to interference among them. Different approaches have been used to reduce or eliminate this interference;
\cite{motion_kinect} applies a small amount of motion to a subset of structured light sensors, while \cite{kinect_time_mux} uses time multiplexing. However, side effects such as motion blur or reduced frame rate prevent its use in FVV applications.

Passive depth devices capture several viewpoints simultaneously and estimate depth by means of algorithms founded in multiview pixel correspondences. Stereo vision has been studied for decades~\cite{scharstein2002taxonomy}, and multiple real-time stereo depth devices are commercially available~\cite{zed}\cite{realsense_d435i}. Multiview stereo (MVS)~\cite{furukawa2015multi} algorithms are based on the same principle, but benefit from a larger set of images from calibrated cameras to improve the estimation of the surface of 3D objects, at the cost of a high computational complexity that prevents real time operation. Plenoptic cameras~\cite{wu2017light} capture a two-dimensional array of viewpoints with very short baseline, which can be used to estimate depth. However, the commercial availability of plenotic video cameras is limited, and their price is similar to that of high-end professional video cameras~\cite{Raytrix}. The main advantage of passive devices for a multi-camera setting is that passive depth sensors do not interfere with each other, which makes them specially well suited for FVV systems. Additionally, as they do not rely on the reception of emitted signals, the depth estimation range is not limited by the device emission power, and do not require controlled lighting conditions, so they can work outdoors. However, the accuracy of depth estimation is affected by non-Lambertian surfaces, homogeneous textures, and object disocclusions, i.e. pixels at the objects contours are only seen by one camera of the stereo pair, hindering the estimation of stereo correspondences in those regions. Stereo disocclusions typically generate holes in the depth map, which can be mitigated by inpainting methods~\cite{jbf_depth}.

\begin{figure*}[t]
\centering{}\includegraphics[width=0.95\linewidth]{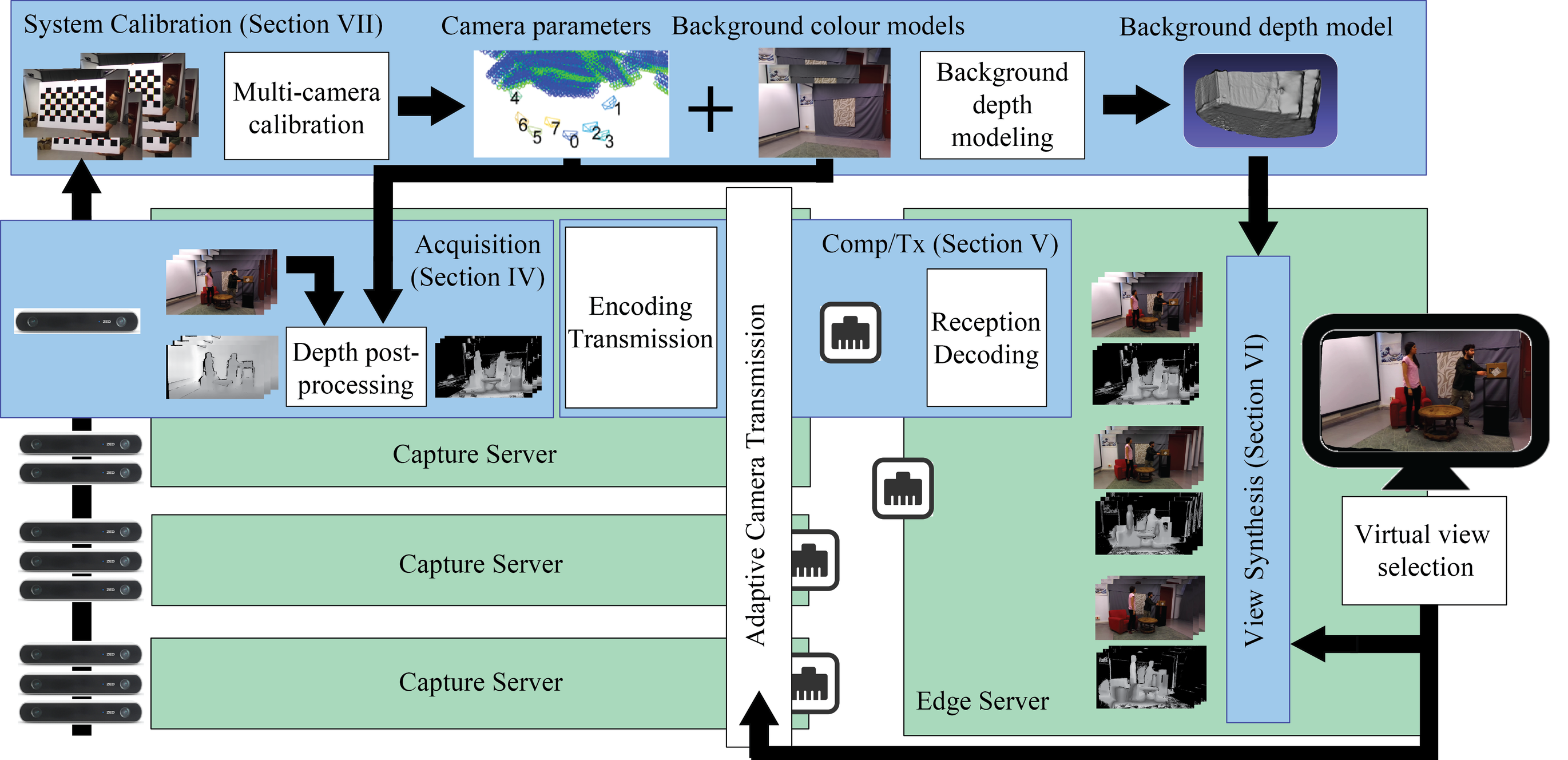}\caption{\label{fig:system_scheme}Block scheme of the \SystemName\ system. Blue boxes indicate the functional blocks of the system: acquisition, compression/transmission, view synthesis and system calibration. Green boxes correspond to the CSs and ES, and indicate where these functional blocks are physically executed.}
\end{figure*}

\subsection{Compression and transmission}
\noindent The delivery of FVV data over existing networks presents a major challenge with respect to 2D video, as it requires the transmission of multiple views plus scene geometry information. Video volumes such as the ones captured by acquisition systems like 4DReplay and Intel True View are several tens bigger than 2D video. Estimations for current and future six-degrees-of-freedom (6DoF) applications suggest that their bitrate requirements are over two orders of magnitude higher than the capacity of broadband internet~\cite{hinds_icme17}. The reduction of the bitrate requirements for FVV applications has been tackled by two different and complementary strategies.

First, substantial bitrate savings can be achieved by reducing the number of transmitted views for a given free-view angle, but this necessarily requires the capacity to render skipped views at the receiver. Also, the reduction in camera density can compromise the quality of virtual intermediate views. Second, efforts have been invested in improving the compression performance of standard video codecs for multiview video, by exploiting inter-camera redundancies. Multiview versions of standard video codecs (such as H.264/AVC or H.265/HEVC), can save up to 30\% bitrate for dense multiview arrays compared to simulcast~\cite{overview_3DHEVC}, at the cost of increasing the encoder complexity. These savings are however considerably reduced for sparse multiview arrays. Additionally, the transmission of a compressed depth sequence per view increases the bitrate around 30\% compared to multiview colour video~\cite{chapter_fvv}.

Regarding transmission, as for conventional 2D video, the  media transport technology (including protocol, strategy, etc.) completely relies on the objectives and characteristics of the application~\cite{schierl2011transport}. For real-time systems like \SystemName, compressed video is commonly transmitted using the RTP/UDP combo. In particular, typically, an independent RTP session is established for each type of data and view.

\subsection{View synthesis}
\label{sota-view-synthesis}

\noindent Virtual view rendering in FVV systems is an extension of the multiview 3D reconstruction problem~\cite{mvs_review} to the temporal dimension. Due to the interest of FVV systems for immersive telepresence applications~\cite{microsoft_holoportation}\cite{mekuria17}, 3D reconstruction of human shapes has been, since early FVV systems~\cite{Carranza:2003:FVH}, very prevalent in the literature~\cite{PAGES2018192}\cite{liu_10}.

Smolic~\cite{SMOLIC20111958} reviews a  continuum of virtual view rendering techniques, and their relation to the different 3D representation formats. Due to the availability of real-time passive depth devices, depth-image-based rendering (DIBR) techniques~\cite{DIBR_Fehn}, based on the 3D warping principle~\cite{3d_warping}, have been widely used in FVV. However, inherent characteristics of MVD data hinder the feasibility of high-quality DIBR view synthesis. The discrete nature of depth pixels (both in image coordinates and depth range) results in ``crack'' artifacts in the virtual view~\cite{DIBR_hole_filling}, which add to the disocclusion problem of 3D warping due to the fundamental limitations of stereo-based depth estimation mentioned in Section~\ref{subsec:capture_sota}. These issues have made DIBR go hand in hand with hole-filling techniques, be it through inpainting~\cite{DIBR_hole_filling} or the integration of the contribution of multiple cameras in virtual view synthesis~\cite{Dziembowski16}. Unfortunately, the use of several reference cameras introduces problems of its own, as errors in camera calibration and depth measures create mismatches in the fusion of images warped from multiple physical cameras. This can cause distortions like double imaging. Naturally, the influence of all these unwanted effects in the quality of the virtual view is more prominent for sparser camera settings.

Some works have constrained the rendering problem, either to improve the synthesis quality or to alleviate the computational cost, using prior knowledge of the scene/BG appearance and/or geometry. This approach has been common in videoconferencing and sport applications. Carranza et al.~\cite{Carranza:2003:FVH} already used pose estimation and BG subtraction to perform real time synthesis of human shapes. The work in~\cite{Suenaga15} describes a FVV system for soccer, in which a model of the field is used to perform BG subtraction and players are inserted in the virtual view by means of billboard rendering. Also for sports,~\cite{fast_plane_based_vs} targets real-time view synthesis combining BG subtraction and projection of the FG masks to the virtual view using per-plane homographies. 

Other works do not require prior knowledge of the scene/BG geometry, but take advantage of a BG appearence model (either static or dynamically learnt) to mitigate disocclussion problems in view synthesis. The work in~\cite{Luo_BG_FG_rendering} provides a review of BG-aware virtual view synthesis methods and inpainting techniques, and proposes its own FG removal method to deal with holes in the virtual view. In the same spirit,~\cite{Mustafa_2016_CVPR} uses multiview object segmentation and temporal coherence between frames to improve the virtual view rendering.

\section{System Overview}
\label{sec:system_overview}

\noindent Fig.~\ref{fig:system_scheme} shows the block scheme of the \SystemName\ system. The following sections describe the real-time functional blocks that compose the system. Acquisition (Section~\ref{sec:capture}), is based on multiple stereo cameras that yield MVD data, where each depth stream is post-processed to improve depth measurements and remove BG data to reduce the bitrate. In the  compression/transmission stage (Section~\ref{sec:compression_and_transmission}) each color/depth data is encoded and transmitted independently using standard video codecs, using an strategy to avoid depth coding distortion. The view synthesis (Section~\ref{sec:rendering}) is designed to generate high-quality virtual views, using a layered approach that takes advantage of BG geometry modeling without the loss of generalization of predefined BG geometries. 

The offline calibration of the system is described in Section~\ref{sec:system_calibration}: it includes the calibration of the multi-camera set and the modeling of the geometry of the static elements of the scene (BG). This is a one-time process that is performed in the case of the reconfiguration of the cameras or a relevant change in the scenario.

Fig.~\ref{fig:system_scheme} also shows how these functional blocks are distributed among physical servers: several CSs at the acquisition side, and one ES at the rendering side. The results in this paper correspond to a system with three CSs handling a total of nine cameras, but the system allows scalability for an arbitrary number of cameras and CSs. Further details on the implementation of the system (servers and network) can be found in Section~\ref{sec:implementaion}.

\section{Acquisition}
\label{sec:capture}

 \begin{figure}[t]
\centering{}\includegraphics[width=\columnwidth]{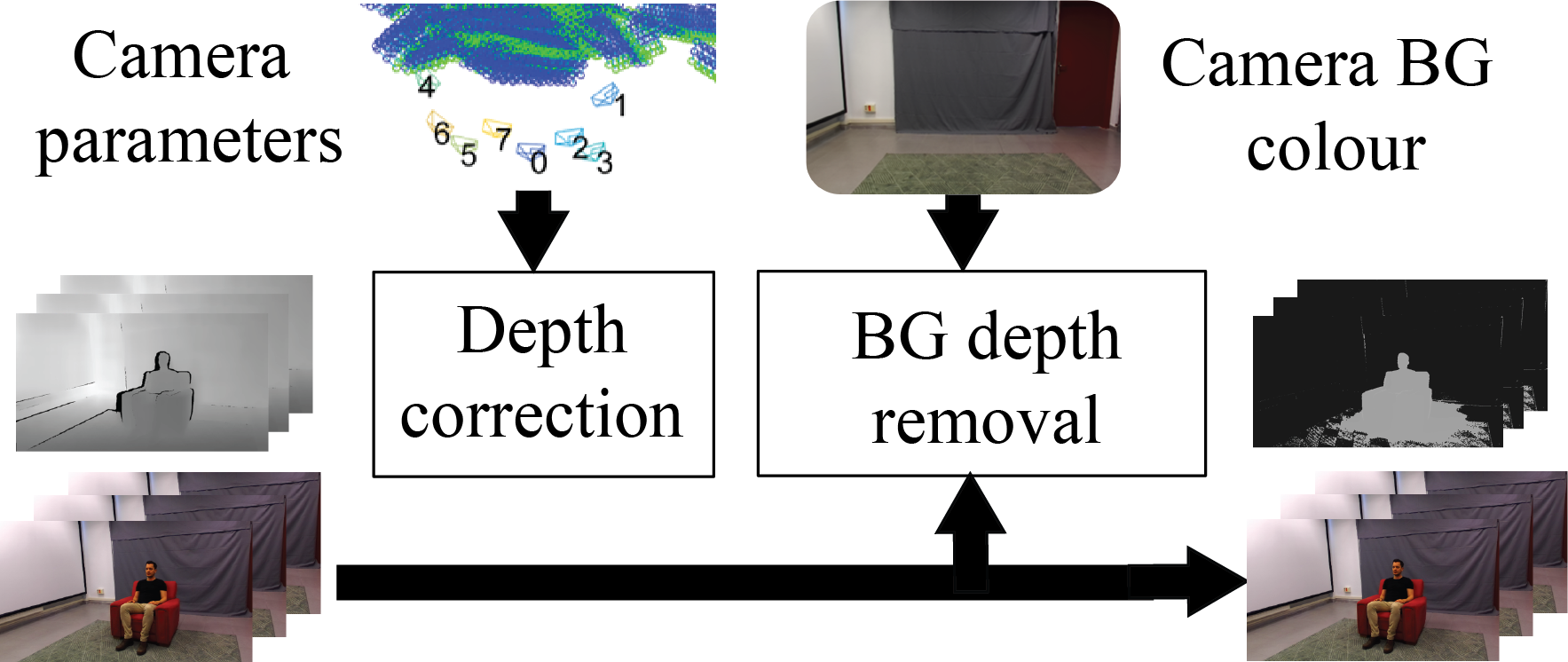}\caption{\label{fig:capture_scheme} Scheme of the acquisition block of \SystemName\ for one camera}
\end{figure}

\noindent Fig.~\ref{fig:capture_scheme} shows the processing scheme of the acquisition block for the two data streams yielded by one camera: colour and depth. The system has been implemented using passive stereo cameras (specifically Stereolab ZED~\cite{zed}) to avoid interference between multiple active devices. The depth data is processed by two blocks: (i) a depth correction block that fixes the depth estimation errors caused by those from the calibration of stereo pairs, and (ii) a BG depth removal block. The colour data is directly passed to the compression/transmission block.

\subsection{Depth correction} 
\label{subsec_depth_post}

\begin{figure}[t]
\centering
\includegraphics[width=\columnwidth]{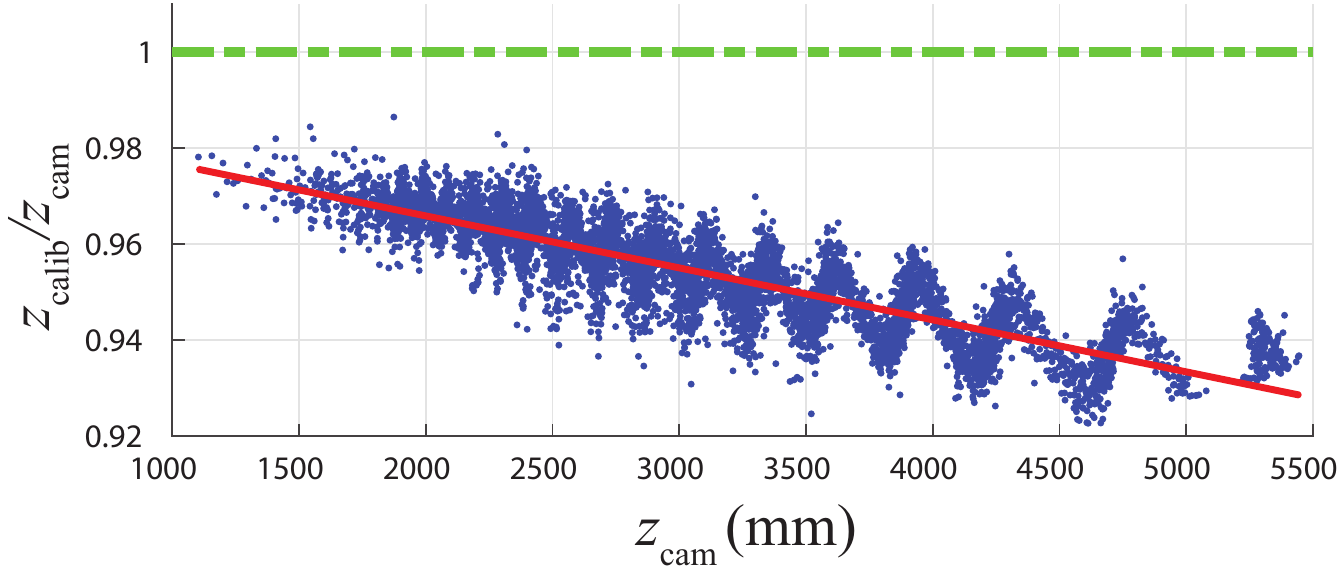}%
\caption{\label{fig:depth_zed_analysis} Depth error model for a single ZED camera. $z_{\textrm{calib}}$ and $z_{\textrm{cam}}$ are, respectively, robust multi-camera and single-camera estimates of the same depth values (data obtained at the multi-camera calibration stage). The green dashed line indicates the ideal relation between both estimates in the absence of error. The red line models the systematic error due to a stereo pair calibration error.}
\end{figure}

\begin{figure}[t]
\centering
\subfloat{\includegraphics[width=0.40\columnwidth]{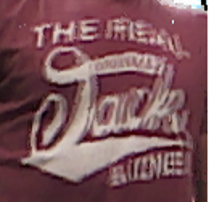}}
\subfloat{\includegraphics[width=0.40\columnwidth]{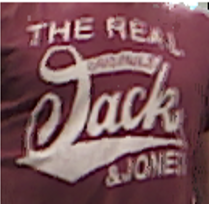}}\\ \vspace{-3.5mm}
\caption{\label{fig:Double-image-artifacts}Example of double-image artifacts in virtual views corrected by the depth correction block. Left: detail of virtual views obtained from uncorrected depth data. Right: detail of virtual views obtained from corrected depth data}
\end{figure}

\noindent The depth processing pipeline includes a block to correct the depth estimation errors caused by slight errors in the calibration of the stereo pair (typical in consumer depth cameras). While slight stereo-pair calibration errors may not be critical in several single-camera applications, it causes an unacceptable degradation in views rendered from MVD data. This block leverages on the data obtained from multi-camera calibration to estimate the calibration error of each stereo-pair and correct depth values independently in each camera. Such block is specially useful in FVV systems using consumer-grade depth cameras, as depth measures can be corrected without accessing the calibration of the stereo pair (not always accessible).

Depth measures obtained by stereo-matching algorithms typically present errors (or noise) that can stem from different sources such as: (i) a stereo pair calibration error, (ii) stereo matching errors, e.g. homogeneous textures, or (iii) the discrete nature of digital images. In particular, errors in the calibration of rectified stereo pairs result in a depth measure error that increase non-linearly with the distance of objects to the camera~\cite{zou_depth_error}.

Fig.~\ref{fig:depth_zed_analysis} illustrates the analysis of the error in depth measures provided by a ZED camera. A further analysis on the depth error model for this device can be found in~\cite{zed_error_model}. The figure compares robust depth estimates ($z_{\textrm{calib}}$), obtained from multi-camera calibration, with the corresponding depth values estimated by a single ZED camera ($z_{\textrm{cam}}$). The data is presented by the ratio $\nicefrac{z_{\textrm{calib}}}{z_{\textrm{cam}}}$ plotted against $z_{\textrm{cam}}$. In the absence of errors in $z_{\textrm{cam}}$,  $z_{\textrm{calib}}=z_{\textrm{cam}}$, and thus  $\nicefrac{z_{\textrm{calib}}}{z_{\textrm{cam}}}=1$ for all $z_{\textrm{cam}}$ values (green dashed line in Fig.~\ref{fig:depth_zed_analysis}). The stereo pair calibration error is manifested as the systematic deviation of the data points from this ideal line (increasing with distance). The ``zig-zag'' pattern responds to errors due to pixel discretization, filtered by smoothing algorithms for depth estimates in continuous surfaces.

All depth measure errors degrade the quality of synthesized images, but the calibration error results in double-image artifacts in the synthesis of a virtual view from MVD data, i.e. multiple reference cameras with unpaired calibration errors contribute to the synthesized image. An example of such artifacts is shown in Fig.~\ref{fig:Double-image-artifacts}.

The depth correction block uses a quadratic model to correct the effect of the calibration error in the depth values, solving these double-image artifacts in the synthesized views. Also, a quadratic error model was proposed in~\cite{kinect_error} for Kinect depth data. A corrected depth value $z_{\textrm{post}}$ is obtained for each pixel of the depth map, as follows: 
\begin{equation}
    z_{\textrm{post}} = (\alpha \times z_{\textrm{cam}} + \beta) \times z_{\textrm{cam}}.
\end{equation}

Parameters $\alpha$ and $\beta$ are estimated from $z_{\textrm{calib}}$ values (and corresponding $z_{\textrm{cam}}$ values) obtained at multi-camera calibration. Multi-camera calibration provides the 3D position of cameras and control points (typically features of a moving checkerboard pattern or similar). This data provides robust estimates for the depth values of those control points from each camera ($z_{\textrm{calib}}$). Synchronously, the depth value of the same control points can be estimated from each camera ($z_{\textrm{cam}}$) obtaining a dataset that captures the accuracy of each depth camera, such as the one in Fig.~\ref{fig:depth_zed_analysis}. $\alpha$ and $\beta$ correspond to the slope and intercept of a linear regression model estimated for the data in Fig.~\ref{fig:depth_zed_analysis} (red line).

\begin{figure*}[t]
\centering
\includegraphics[width=0.95\linewidth]{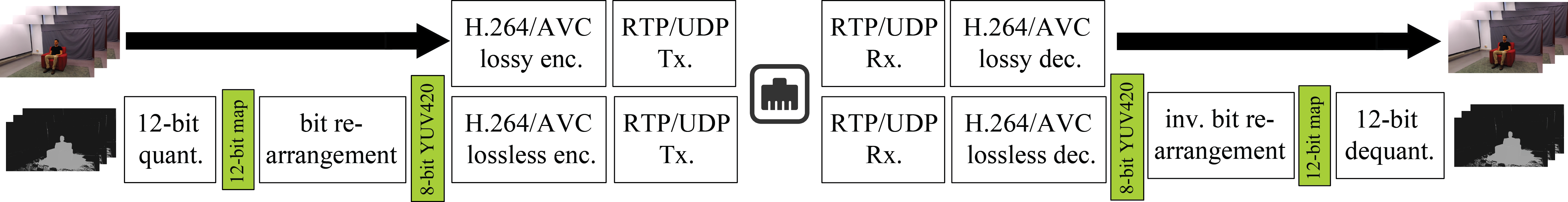}
\caption{Scheme for the compression and transmission of the colour and depth streams of one camera in the \SystemName\ system}
\label{fig:comp_transm_arch}
\end{figure*}

\subsection{Background depth removal}
\label{sec:segmentation}

\noindent Given that the camera set of the \SystemName\ system is static, it can be assumed that, in most use-cases, a considerable region of the scene captured by each camera corresponds to a static BG. Unfortunately, as BG textures are often homogeneous, accurate BG depth measurements from stereo-matching algorithms are not possible, resulting in wrong and highly time-varying values. Thus, the BG depth removal block serves a double purpose. First, the continuous transmission of BG depth values can be spared, resulting in relevant bitrate savings. Second, depth values for BG pixels can be pre-computed at the system calibration, applying more accurate techniques than real-time depth estimation (see Section~\ref{sec:system_calibration}). This translates into higher quality virtual views.      

A BG colour model is used to classify FG/BG pixels in the colour frame. However, only the depth values of BG pixels are skipped from transmission (only the BG geometry). The BG colour data is transmitted to allow variations in the appearance of objects due to illumination changes, for example. We use a classical single gaussian model~\cite{wren1997pfinder,gallego_mahalanobis}, which is computationally cheap and good enough for static backgrounds.

\section{Compression and transmission}
\label{sec:compression_and_transmission} 

\noindent Fig.~\ref{fig:comp_transm_arch} shows the coding and transmission scheme for one camera. The colour and depth streams yielded by each camera are encoded and transmitted independently. On the one hand, standard lossy H.264/AVC encoders are used for the colour streams. On the other hand, the coding scheme for depth data is designed to minimize the coding distortion on depth data, as it has a greater impact on the quality of synthesized views. The proposed scheme uses 12-bit depth maps, which are efficiently packed, encoded and transmitted using 8-bit lossless H.264/AVC codecs. The transmitted streams are received and decoded in the ES, to feed the virtual view synthesis block, as shown in Fig.~\ref{fig:system_scheme}. To enable real time operation, GPU-based encoders and decoders have been used.

An adaptive scheme for the transmission of a subset of cameras has been implemented (see Fig.~\ref{fig:system_scheme}). This strategy enables bitrate savings and allows the scalability of the system, i.e. allows to increase the number of cameras without increasing the network capacity.

\subsection{Coding scheme for depth data}
\label{subsec:depth_coding_scheme}

\noindent The depth coding scheme has been designed to preserve the precision of depth data using off-the-shelf GPU-based video codecs, generally implemented for 8-bit YUV420 video input. Therefore, since CSs generate 16-bit depth maps, a further quantization process is needed to meet the encoder requirements. Nevertheless, given that depth maps are single-band signals, the 4 chroma-bits per pixel of the YUV420 format can be used to extend depth data representation from 8 to 12 bits. These extra bits for depth data representation avoid most synthesis artifacts related with over-quantization of depth data ("cracks" in the synthesized image). Therefore, prior to the 8-bit lossless encoder, the 16-bit depth maps are processed by two blocks: (i) 12-bit quantization and (ii) a bit re-arrangement block that maps the 12-bit depth maps onto 8-bit YUV420 images. This bit re-arrangement is based on the work by Pece et al.~\cite{pece2011adapting}, but designed specifically for a lossless YUV420 codec, allowing to transmit 12-bit depth maps.

At the decoder side, the inverse quantization and bit-rearrangement operations are performed to recover the original depth information.

\begin{figure*}[t]
\centering
\includegraphics[width=0.70\linewidth]{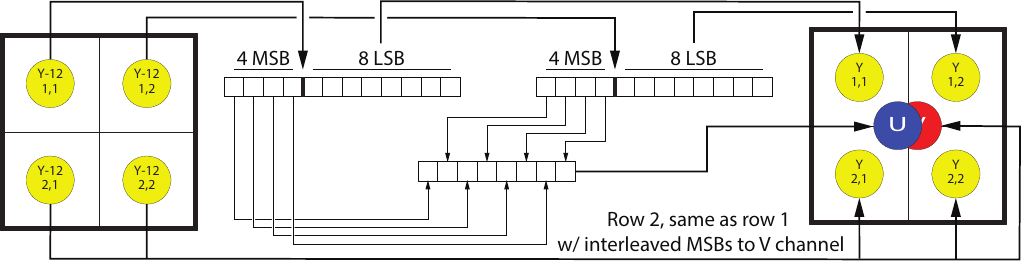}
\caption{\label{fig:depth_coding_scheme}Bit distribution of 4 pixels (2x2 block) of a
12-bit disparity map (Y-12) onto a 2x2 block of a 8-bit YUV420 structure}
\end{figure*}

\begin{figure}[t]
\centering
\subfloat[]{\includegraphics[width=0.33\columnwidth]{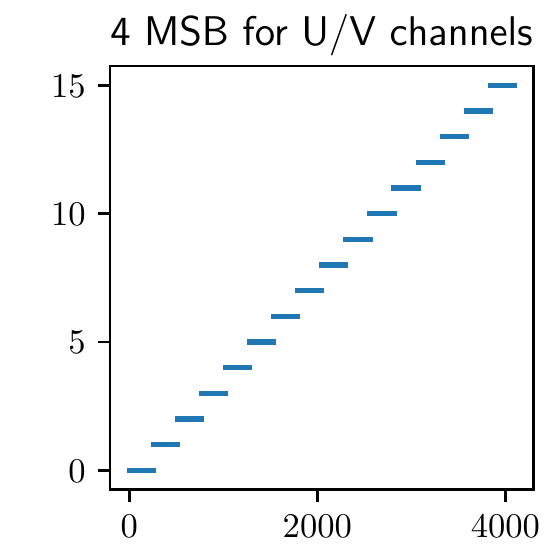}}
\subfloat[]{\includegraphics[width=0.33\columnwidth]{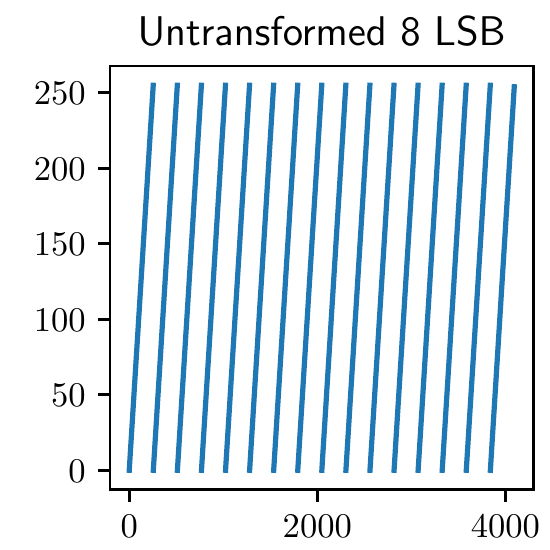}}
\subfloat[]{\includegraphics[width=0.33\columnwidth]{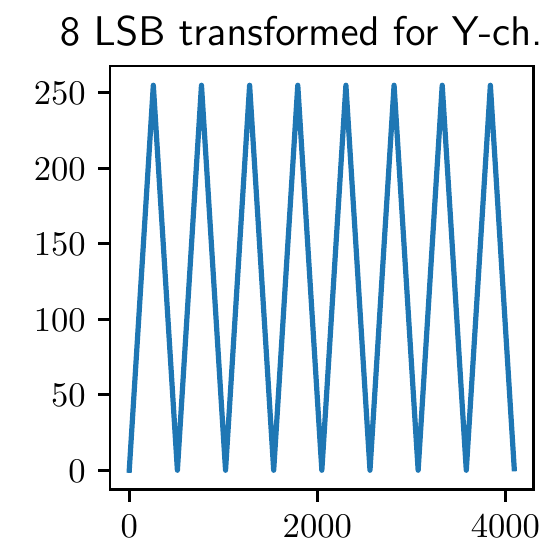}}
\caption{\label{fig:depth_decomposition} Values of the 4 MSBs and 8 LSBs of a 12-bit number. (a) Evolution of the 4-MSB value (y-axis) for an increasing 12-bit number (x-axis); (b) evolution of the 8-LSB value for an increasing 12-bit number; and (c) evolution of the 8-LSB value for an increasing 12-bit number if~(\ref{eq:3}) is applied. Note that the discontinuities in (b) are removed in (c), increasing the correlation of the 8 MSBs extracted from contiguous 12-bit values.}
\end{figure}

\subsubsection{12-bit quantization}
\label{12 bit quantization}

A common non-linear quantization is used to map depth values to a 12-bit disparity map~\cite{CARBALLEIRA2016}. Additionally, the 0-disparity value is reserved to signal BG pixels. FG depth values are clipped within the [$z\textsubscript{near}$,$z\textsubscript{far}$] range (nearest and farthest clipping planes). The 1-disparity value is reserved to signal values out of this range (including pixels with no depth value, e.g. object boundaries). 
Thus, FG depth values in range ($z$) are mapped to disparity values ($Y$) using the following expression: 
\begin{equation}
Y=\left(2^{12}-3\right)\left(\dfrac{1}{z}-\dfrac{1}{z_{\text{far}}}\right)/\left(\dfrac{1}{z_{\text{near}}}-\dfrac{1}{z_{\text{far}}}\right)+2
\label{eq:2}
\end{equation}
Signaling BG pixels with a reserved depth value allows the synthesis module to identify them to perform the layered view synthesis while saving bitrate (BG regions do not increase the number of bits of the coded depth map).

\subsubsection{Bit re-arrangement}

The 12-bit disparity map is then mapped to the luminance and chrominance channels of an 8-bit YUV420 structure that feeds the video codec. A scheme of this mapping is depicted in Fig.~\ref{fig:depth_coding_scheme}. Each 2x2 pixel block of the 12-bit map is mapped to a 2x2 pixel block of the YUV420 structure as follows: 
\begin{itemize}
    \item The 8 least significant bits (LSBs) of each 12-bit pixel are mapped to the 8-bit Y value of the corresponding YUV420 pixel.
    \item The 4 most significant bits (MSBs) of the 4 12-bit pixels are mapped to the U,V values of the 2x2 YUV420 block. 
\end{itemize}

To increase the correlation of neighbour pixels in the YUV420 and, therefore, the coding efficiency, MSBs and LSBs are re-ordered in the mapping process as follows:
\begin{itemize}
    \item The 4~MSBs of pixels (1,1) and (1,2) are bit-interleaved as shown in Fig.~\ref{fig:depth_coding_scheme} and mapped to the U value. Analogously, the 4~MSB of pixels (2,1) and (2,2) are bit-interleaved and mapped to the V value.
    \item  The 8~LSBs are mapped to Y values according to~(\ref{eq:3}).
\end{itemize}
\begin{equation}
\text{Y channel}=\begin{cases}
8~\text{LSB} & \text{, if 4~MSB mod 2}=0\\
255-8~\text{LSB} & \text{, if 4~MSB mod 2}\neq0
\end{cases}\label{eq:3}
\end{equation}

Bit interleaving of the 4-MSBs of pixel pairs to U,V channels maps the bits with higher variability to the least significant positions of the U,V values, increasing correlation of adjacent U,V values in the YUV420 structure. With the same target, the mapping strategy for 8-LSBs avoids the discontinuities in the values of the 8~LSBs of contiguous 12-bit numbers (it is equivalent to the modulo-256 operation). Fig.~\ref{fig:depth_decomposition} shows how these discontinuities disappear if~(\ref{eq:3}) is applied.

\subsection{Transmission}
\noindent Encoded streams are transmitted independently over RTP/UDP to allow real-time operation. RTP packets are conformed to have a maximum size equal to the Maximum Transmission Unit (MTU) to avoid IP fragmentation. To that end, we apply the application-level fragmentation scheme included in RFC3984~\cite{rfc3984}. At the decoder side, received RTP packets are reassembled to conform complete NAL Units. Frame timestamps are obtained from the CS clocks, which are synchronized using the Precision Time Protocol (PTP)~\cite{ptp}, and transmitted to the ES over RTP.

\subsection{Adaptive/Dynamic view transmission}
\noindent For the cases of a limited number of clients, it is possible to reduce significantly the transmitted bitrate, as only the streams from a subset of cameras are required at a given moment. This subset is therefore selected dynamically, depending on the viewpoint/s selected by the client/s. Specifically, $N_{\text{Tx}}$ of the $N$ cameras are transmitted (those closest cameras to the virtual viewpoint). The  computational load of CSs and the transmission bitrate are reduced approximately in the same proportion. The indices of the required $N_{\text{Tx}}$ cameras are transmitted from the ES to the CSs using MPI messages~\cite{MPI}.

In our implementation, with one client, $N_{\text{Tx}}$ is set to five cameras. While the rendering block uses three reference cameras to synthesize the virtual view (see Section~\ref{sec:rendering}), two additional views are transmitted to guarantee that the rendering block has, at any time, all necessary reference views, even in the case of a rapid movement of the virtual viewpoint. 

\section{View synthesis}
\label{sec:rendering}
\noindent The synthesis module is fundamentally based on the DIBR paradigm~\cite{DIBR_Fehn,3d_warping},
which produces synthesized views from viewpoints other than those of
the reference cameras from the colour and depth information of the latter. The basic idea of DIBR warping is simple:
each pixel in an image from a calibrated camera corresponds to a line
in 3D space, and depth information constrains
that line to a single point, establishing a correspondence between
pixels and 3D points. Thus, each reference image encodes
a 3D point cloud that can be projected on a virtual camera, placed
wherever the user chooses.

As already mentioned in Section~\ref{sota-view-synthesis}, the discrete nature of images causes "crack" artifacts whenever an object covers more pixels in the synthesized view
than in the reference view. This problem is handled using backward
warping~\cite{hornung2009,berjon2011backwardmapping}, combined
with inpainting or image filtering techniques. The other fundamental
problem of DIBR is the \emph{disocclusion problem}, whenever the virtual view peeks behind an object, uncovering a part of the scene
not visible in the reference image. The only solution to this limitation
is to provide multiple reference cameras that cover all
relevant surfaces of the scene to be synthesized.

Mixing information from multiple reference cameras is trivial for perfect depth data~\cite{3d_warping}, but poses more of a significant problem the noisier
the input data is. Unfortunately, depth estimations from real-time
stereo matching are inherently noisy, especially in uniform BGs, which
makes it impossible to reconcile inputs. Observing that the defining
characteristic of the BG is that it is stationary, we can produce a detailed off-line geometry model of the BF during system
calibration (see section \ref{sec:system_calibration}). This results in MVD data of the BG which is consistent among all
cameras. During online operation of the system, only depth information of FG objects needs to be transmitted (see Sections \ref{sec:segmentation} and \ref{subsec:depth_coding_scheme})  

\begin{figure}[t]
\begin{centering}
\includegraphics[width=1\columnwidth]{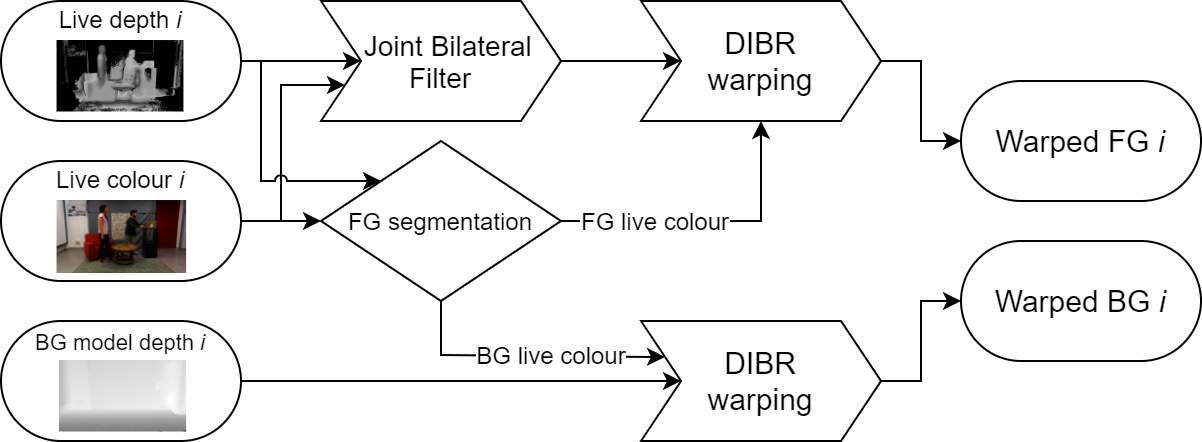}
\par\end{centering}
\caption{\label{fig:Per-camera-process-to}Per-camera process to generate contributions to the FG and BG layers.}
\end{figure}

To generate the synthesized view, we first obtain per-camera contributions
to the FG and BG layers as detailed in Fig.~\ref{fig:Per-camera-process-to}.
Live depth data is pre-processed to close small holes
and gaps around the edges of objects using a joint bilateral filter~\cite{camplani2013}. FG and BG areas of the live colour are separated using the signalling in depth information. Live FG depth and colour data is used to warp the FG to the virtual camera. The BG
is warped using the depth information from the static model, but we use the BG live colour rather than the static BG colour because moving objects
cast shadows and reflect diffuse light onto the scene; hence, the
appearance of the synthesized view is more natural when using live colour.

\begin{figure}[t]
\begin{centering}
\includegraphics[width=1\columnwidth]{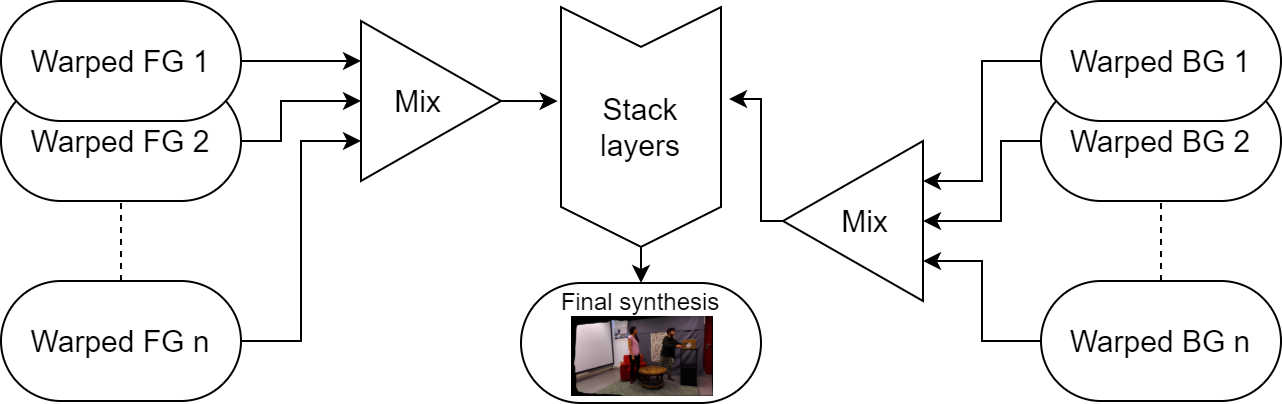}
\par\end{centering}
\caption{\label{fig:mixing-and-stacking-fg-bg}FG and BG contributions
from the reference cameras are mixed separately and finally composited
to yield the final virtual image.}
\end{figure}

Applying this process to all active cameras (typically three for real-time performance constraints), we end up with multiple contributions to the warped FG and BG. As shown
in Fig.~\ref{fig:mixing-and-stacking-fg-bg}, all the contributions
to the FG and BG layers are mixed together, according to the distance
of each reference camera to the virtual camera. Finally, the
combined FG and BG layers are stacked to composite
the final image. \SystemName\ synthesis examples are shown in Section~\ref{sec:subjective_evaluation}, Fig.~\ref{fig:Tests-screenshots}.

\section{System calibration}
\label{sec:system_calibration}

\noindent Any FVV system requires a precise multi-camera calibration, as DIBR algorithms rely in MVD data from calibrated cameras.  Several well-known multi-camera calibration algorithms can be used~\cite{hartley_zisserman_2004}, and those that use known calibration objects such as checkerboard patterns provide accurate results. As already mentioned, the 3D positions of control points obtained in calibration are used to correct depth measures in the acquisition block (see Section~\ref{sec:capture}).

Additionally, the calibration of \SystemName\ includes the estimation of a BG colour model (used in BG depth removal) and a BG depth model (used in view synthesis). Using information from all cameras, a 3D reconstruction of the static BG is obtained, which is then re-projected to each camera to obtain consistent per-camera BG depth models that will be used for the layered synthesis (see Section ~\ref{sec:rendering}). The structure from motion (SfM) paradigm~\cite{hartley_zisserman_2004}, is used to recover the 3D structure of the BG from colour images from the set of calibrated cameras. The following steps are used to obtain an accurate 3D reconstruction of the BG: 
\begin{enumerate}
    \item Input images are filtered using a multi-scale Retinex filter~\cite{retinex}, as it has been shown~\cite{berjon2016fastfeaturematching} to dramatically increase the number of extracted feature points and the level of detail of the resulting point cloud.   
    \item Feature points are detected and matched over the multiview set using AKAZE features~\cite{akaze}.
    \item An initial point cloud is triangulated from point matches and known multi-camera calibration.
    \item A MVS variational algorithm based on photo-consistency constraints~\cite{mvs_variational} is used to obtain the final 3D structure.
\end{enumerate}

\section{Implementation details}
\label{sec:implementaion}

\noindent This section describes implementation details and performance tests proving that the \SystemName\ system has been designed and dimensioned to work in real-time using off-the-shelf hardware. As shown in Fig.~\ref{fig:system_scheme}, each CS holds the capture and encoding/transmission of several cameras, while the ES is in charge of the reception/decoding of all streams and view synthesis. The remainder of the section gives details on those servers and network.    

\subsection{Capture Servers}

\begin{table}[t]
\centering
\caption{\label{tab:GPU-encoding-load} GPU load tests in CSs. Each configuration indicates the number of cameras that are assigned to each GPU for depth estimation. Encoding of all streams is assigned to the Encoding GPU in all cases}
\begin{tabular}{c|c}
\multirow{2}{*}{Camera-GPU configuration} & Max. Avg. Frame \\
& processing time (ms)\tabularnewline
\hline 
\textbf{2@Depth GPU + 1@Encoding GPU} & \textbf{19.47}\\
3@Depth GPU + 1@Encoding GPU & 39.09\\
2@Depth GPU + 2@Encoding GPU & 49.43\\
\hline
\hline 
\end{tabular}
\end{table}

\subsubsection{GPU equipment and limits in the number of cameras per server}
While depth estimation is embedded in the chip of some cameras~\cite{realsense}, depth estimation in ZED cameras requires an external GPU. The GPU computational capacity of the CS is therefore shared for depth estimation, depth post-processing and MVD encoding. Each CS is equipped with 40 PCIe lanes dedicated to GPUs and multiple USB 3.0 controllers managing ZED cameras, so it cannot effectively host more than two GPUs. To enable depth computation and video encoding for multiple cameras, each CS is equipped with one NVIDIA GeForce GTX 1080 (which imposes licensing limitations in the number of encoding instances) and one NVIDIA Quadro P4000 (which does not). 

With ZED cameras operating at 1920x1080p and 30~fps, the computational capacity of the GPUs can handle up to three cameras per CS in real time. Depth computation for two cameras is assigned to the GeForce GTX 1080 (Depth GPU hereafter), while depth computation for the third camera and the encoding of the colour and depth streams is assigned to the Quadro P4000 (Encoding GPU hereafter). Table~\ref{tab:GPU-encoding-load} summarizes experimental results on the capacity of a CS to operate with multiple cameras in different configurations. The viability of the configurations has been tested measuring average frame processing times: time elapsed since the camera captures a frame until it is ready for transmission. Average frame processing times are measured for each stream (colour and depth) and the maximum value (worst case) is reported in the table. Processing times must be below 33~ms per frame to enable real time operation.

\subsubsection{Encoder configuration and MVD bitrates}

\begin{table}[t]
\begin{centering}
\caption{\label{tab:Encoding parameters}Coding parameters for colour and depth streams}
\par\end{centering}
\centering{}%
\begin{tabular}{|c|c|c|}
\hline 
 Parameter & Colour stream & Depth stream\\
\hline 
  \hline
    Compression & Lossy AVC & Lossless AVC \\
  \hline
    Rate control & one-pass VBR & one-pass VBR\\
  \hline
    Target bitrate & 5-15 Mbps & -\\
  \hline
    GOP & IPPPP & IPPPP\\ 
 \hline
    IDR period & 30 frames & 30 frames\\
\hline 
\end{tabular}
\end{table}

\begin{table}[t]
\begin{centering}
\caption{\label{tab:depth_bit_rates} Empirical depth bitrates in different scenarios with different ratios of BG pixels}
\par\end{centering}
\centering{}%
\begin{tabular}{|c|c|c|c|c|}
\hline 
\multicolumn{2}{|c|}{Scenario} & Simple & Medium & Complex\tabularnewline
\hline 
\multicolumn{2}{|c|}{Mean ratio of BG pixels (\%)} & 80 & 75 & 57\tabularnewline
\hline 
Depth bit & w/o. BG depth rem..  & {96-112} & {92-114} & {99-121}\\
\cline{2-5}
rate (Mbps) &  w. BG depth rem. & {34-74} & {43-79} & {79-99}\tabularnewline
\hline 
\end{tabular}
\end{table}

Table~\ref{tab:Encoding parameters} summarizes the coding parameters for depth and colour streams in our implementation of the \SystemName\ system. Target bitrates for colour streams between 5 and 15~Mbps  have been selected empirically for a high visual quality (see Section~\ref{sec:subjective_evaluation}). For depth streams, due to the use of lossless encoders, no rate-control can be applied. We have measured that a complete 1920x1080p~@30~fps depth stream requires 90-120~Mbps (see Table~\ref{tab:depth_bit_rates}). The BG depth removal results in bitrate savings that depend on the complexity of the scenario (ratio of BG pixels). Table~\ref{tab:depth_bit_rates} shows empirical ranges for the bitrate required by depth streams depending in the three scenarios used for subjective evaluation (Section~\ref{sec:subjective_evaluation}). Results show depth bitrate savings from 20\% to 66\%.

Bitrate savings due to BG depth removal, plus the adaptive camera transmission strategy, considerably reduce the required bitrate for the MVD data. In the case of our implementation, with nine reference cameras and $N_{\text{Tx}} = 5$, the required bitrate is 250-550~Mbps instead of 1.5~Gbps.

In our implementation, each CS is physically linked with the ES through a point-to-point 1~Gbps Ethernet link.

\subsection{Edge Server}
\noindent The ES is equipped with one NVIDIA GeForce GTX 1080 that handles the decoding of all received MVD streams and the virtual view synthesis. Below, we present results on the capability of the ES to perform the following operations in real time: i)~decode multiple MVD streams, and ii)~synthesize the desired virtual view and render it at a 1920x1080p resolution and 30~fps.

\subsubsection{Decoding times for multiple streams}

We have measured frame decoding times at the ES with the full \SystemName\ system operating with nine cameras at 1920x1080p @30~fps, all transmitted simultaneously (9 colour + 9 depth streams). The average frame decoding time is 16.5~ms, well below the real time limit of 33~ms for 30~fps. This indicates that parallel decoding of multiple streams in a single server does not impose limitations on the real-time operation of the system.    

\subsubsection{View synthesis}
The maximum frame synthesis time for a 1920x1080p resolution and three reference cameras is 30~ms, which makes it possible to operate the system at 30~fps.

\subsection{End-to-end and motion-to-photon delays}

\noindent \SystemName\ presents an end-to-end delay (time span between frame capture and the presentation of the corresponding virtual frame at the ES) of 252~ms. This value allows the use of \SystemName\ for immersive videoconferences as it is well below the one-way delay threshold of 400~ms for bidirectional video communications~\cite{max_delay}. Given the maximum frame synthesis time of 30~ms, the motion-to-photon (MTP) delay (time span between a viewpoint update command and its result) is in the 30-63~ms range (mean value of 47~ms). The worst-case scenario (63~ms) corresponds to whenever a viewpoint update command is received in the ES just after the synthesis of a new frame has started. In this case, the synthesis time of the frame at the new virtual viewpoint (30~ms) will not start until the current frame period is over (33~ms). The referred MTP delay value range allows for a seamless virtual view navigation.

 \section{Assessment of the visual quality of \SystemName}
\label{sec:subjective_evaluation}

\begin{figure*}[t]
\centering
\includegraphics[width=0.9\linewidth]{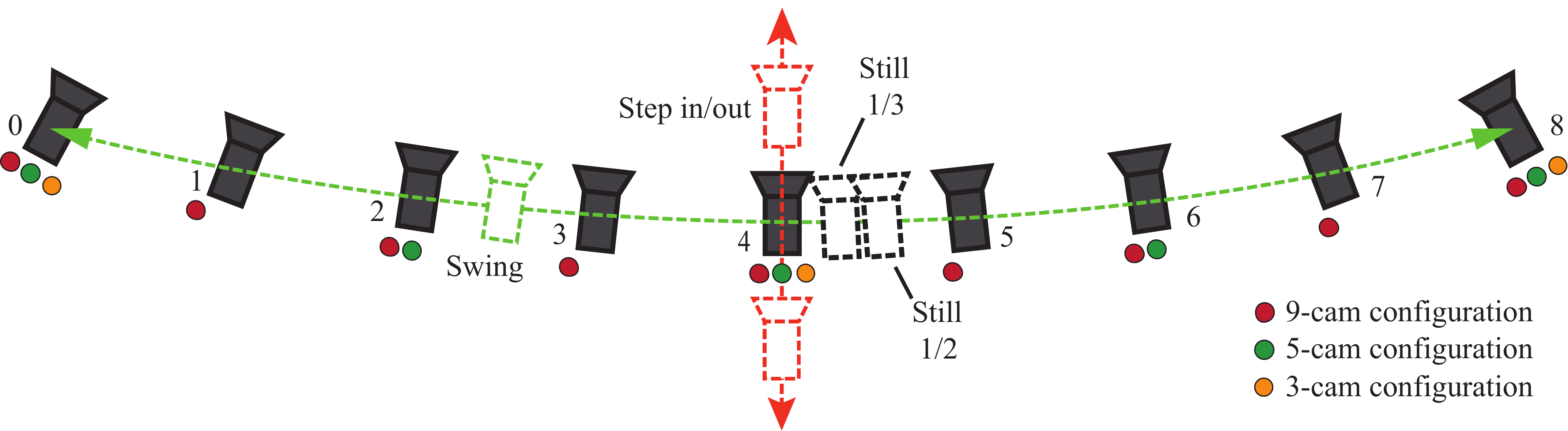}
\caption{\label{fig:camera_setting} Sketch of camera setting for the subjective tests. Black filled cameras represent physical reference cameras. Dashed cameras represent virtual cameras for different trajectories: Green - swing, red - step in/out, black - still cameras at different baseline distances. Colour dots indicate the physical cameras that are used in the three configurations with different camera densities.}
\end{figure*}

\subsection{Introduction}
\noindent The visual quality of \SystemName\ has been assessed by means of a thorough subjective evaluation. Subjective evaluation has been consistently used as the preferred test bench for video quality~\cite{results_FTV} as it directly relies on the feedback of final users. Additionally, in the case of FVV systems, the use of subjective evaluation is particularly necessary due to several reasons:
\begin{itemize}
    \item The absence of reliable objective quality metrics for view synthesis distortions~\cite{Bosc12}.
    \item The lack of reference videos (ground truth) for the virtual viewpoints.
\end{itemize}

Two sets of subjective tests were carried out: (i) a \textbf{comparative quality analysis} between the \SystemName\ synthesis and MPEG's View Synthesis Reference Software (VSRS)~\cite{VSRS}, and (ii) an \textbf{absolute quality analysis}, where the \SystemName\ synthesis is assessed with an absolute quality range. The comparative analysis allows us to compare the \SystemName\ synthesis quality with respect to an state-of-the-art synthesis method. The absolute analysis quantifies the visual quality of the system, taking into account the limitations of the visual quality of the physical cameras of the system. The following subsections describe the test material, environment and equipment, that are common to both analysis, and the particular methodology of each one.

\begin{figure*}[t]
\centering
\subfloat[Simple Scenario\label{fig:vs_simple_scenario}]{\includegraphics[width=0.32\linewidth]{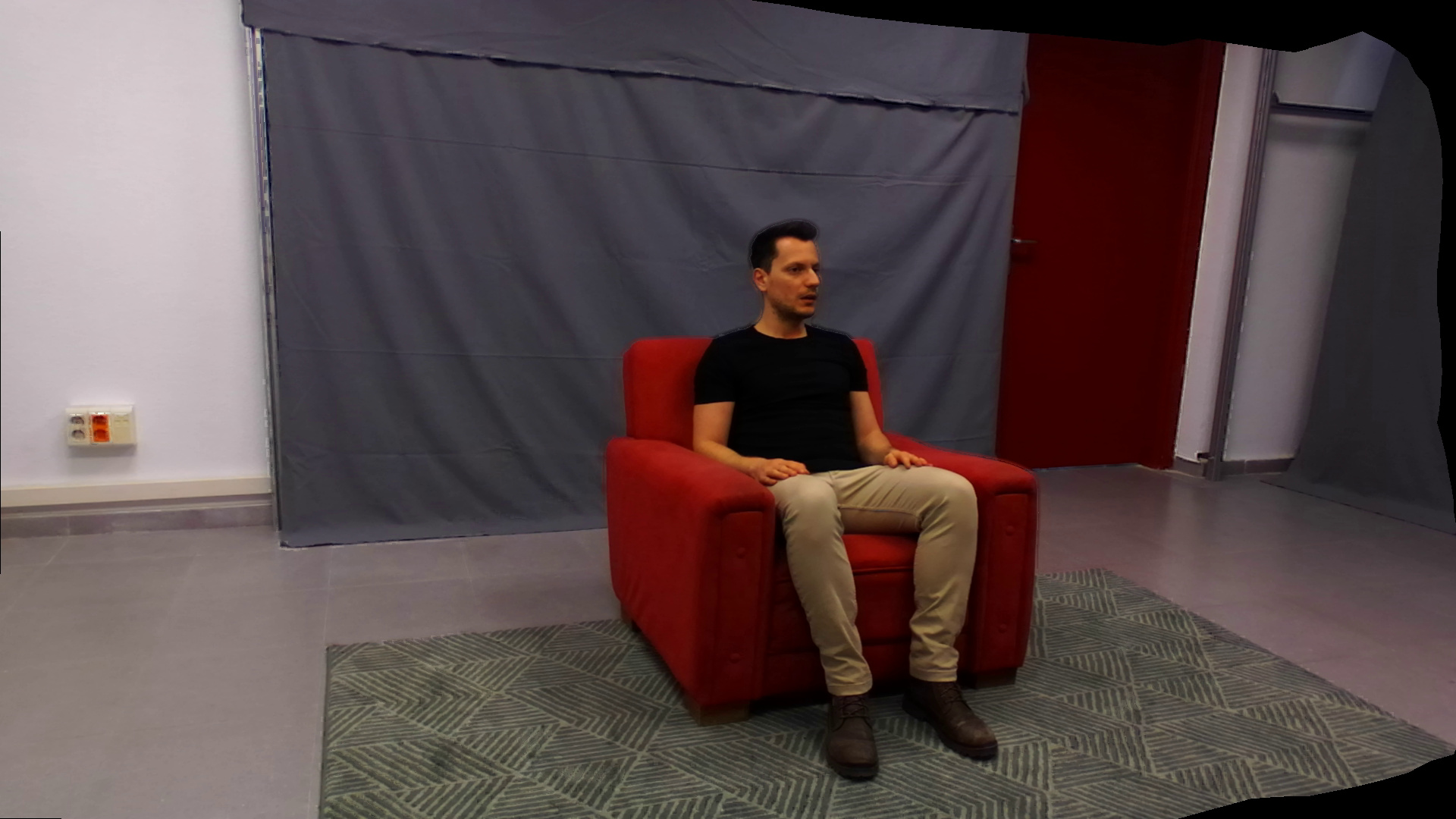}}\hspace{1mm}
\subfloat[Medium Scenario\label{fig:vs_medium_scenario}]{\includegraphics[width=0.32\linewidth]{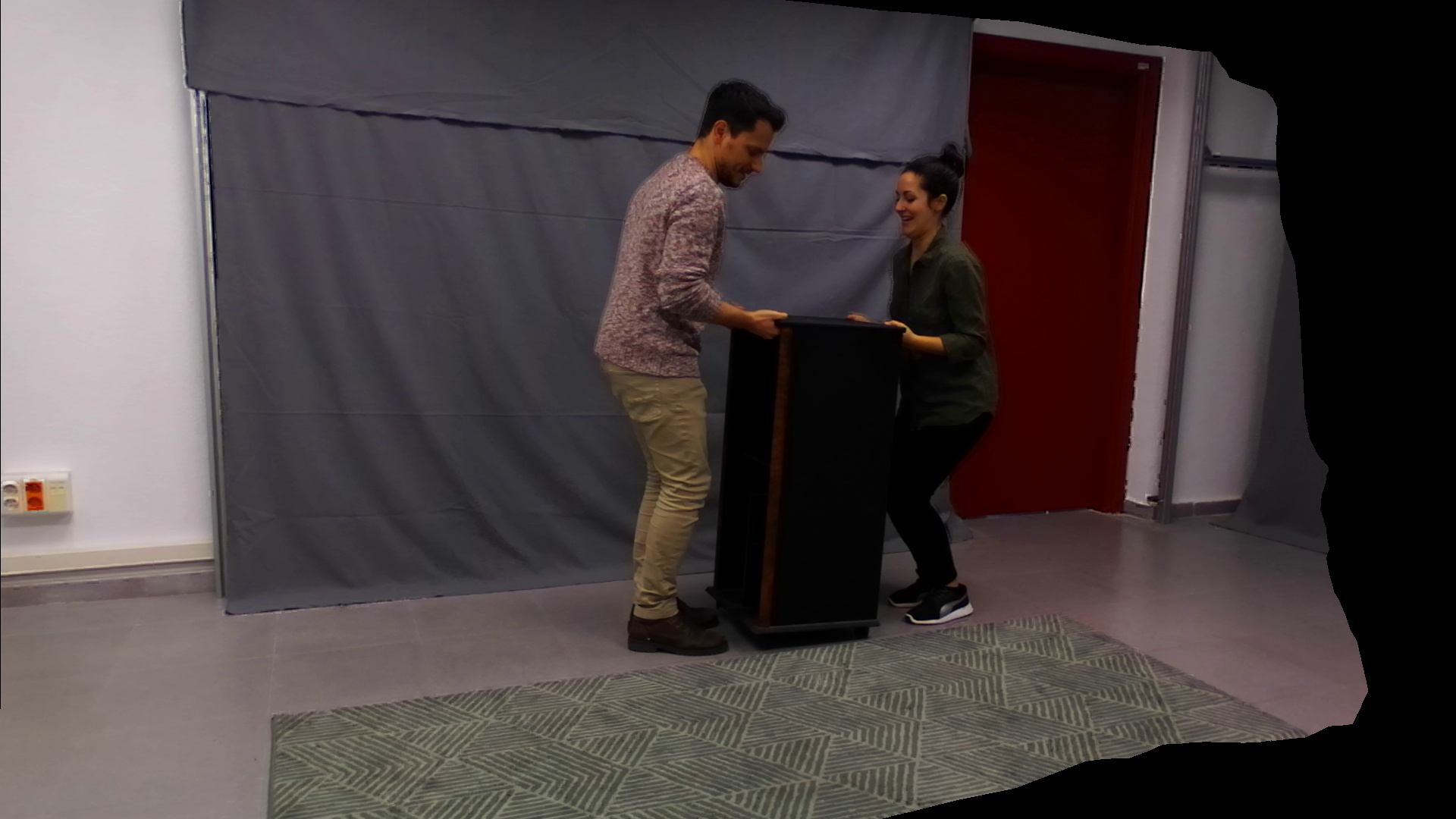}}\hspace{1mm}
\subfloat[Complex Scenario\label{fig:vs_complex_scenario}]{\includegraphics[width=0.32\linewidth]{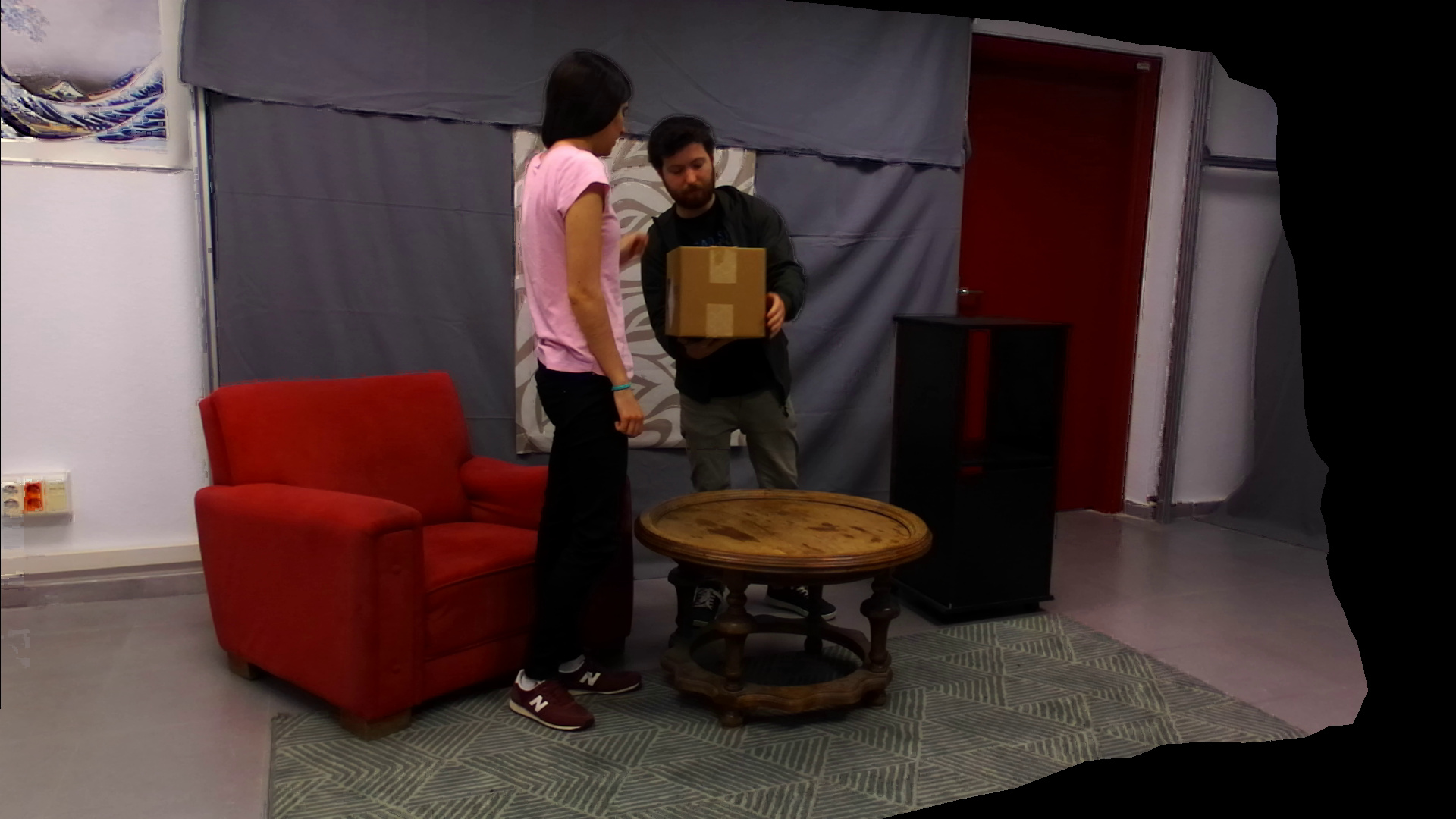}}\\
\caption{\label{fig:Tests-screenshots} Screenshots of test contents for the comparative and absolute quality analyses. The images correspond to synthesized frames  at virtual viewpoints. The black region of the images are regions not captured by the BG model, which are uncovered at that particular virtual viewpoint.}
\end{figure*}

\subsection{Test material}
\noindent The test material consists of 1080p@30fps virtual video sequences, generated using the \SystemName\ and VSRS synthesis methods. These processed video sequences (PVS), as opposed to reference videos (REF) captured by physical cameras, are generated from MVD data captured by a set of nine cameras, disposed in an arch setting, with a separation of 270~mm between them, spanning an angle of 60~degrees. The camera setting is sketched in Fig.\ref{fig:camera_setting}. All PVS are 12 seconds long. Colour sequences were encoded at 15~Mbps. 

The test material includes variability in content, density of the physical camera setting and virtual camera trajectory, so that the competing strategies can be compared under very different conditions:

\subsubsection{Content} Three different types of content (scenarios) were recorded, which entail different levels of complexity, mainly derived from the level of object movement and occlusions. Examples of these scenarios are shown  in  Fig.~\ref{fig:Tests-screenshots}.  
\begin{itemize}
 \item \textbf{Simple scenario}: few objects with little or no movement.
\item \textbf{Medium scenario}: more objects with some movement and some occlusions.
\item \textbf{Complex scenario}: plenty of objects with significant  movement and multiple occlusions.
\end{itemize}

\subsubsection{Camera configuration} Three different camera configurations, with decreasing camera density were used. The camera density is decreased by skipping intermediate physical cameras (increases baseline while maintaining the angle of the virtual camera trajectory, see Fig.~\ref{fig:camera_setting}).
\begin{itemize}
\item \textbf{9-cam configuration}: all 9~cameras are used.
\item \textbf{5-cam configuration}: 5~cameras are used, doubling the original baseline distance.
\item \textbf{3-cam configuration}: 3~cameras are used, quadrupling the original baseline distance.
\end{itemize}

\begin{figure*}[t]
\centering
\subfloat[Average (all cases)\label{fig:PC-general}]{\includegraphics[width=0.49\columnwidth]{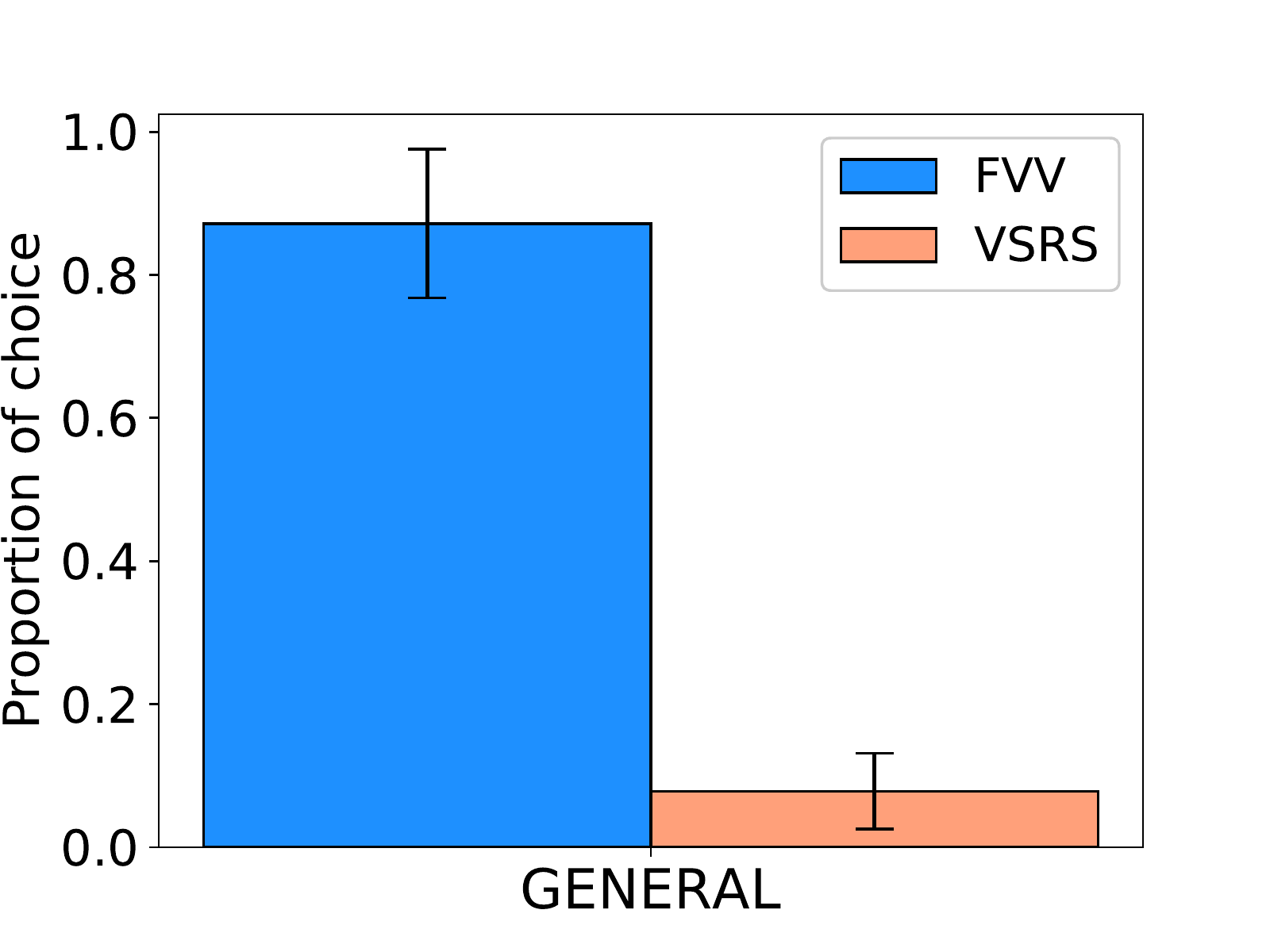}}
\subfloat[Camera configurations\label{fig:PC-ncameras}]{\includegraphics[width=0.49\columnwidth]{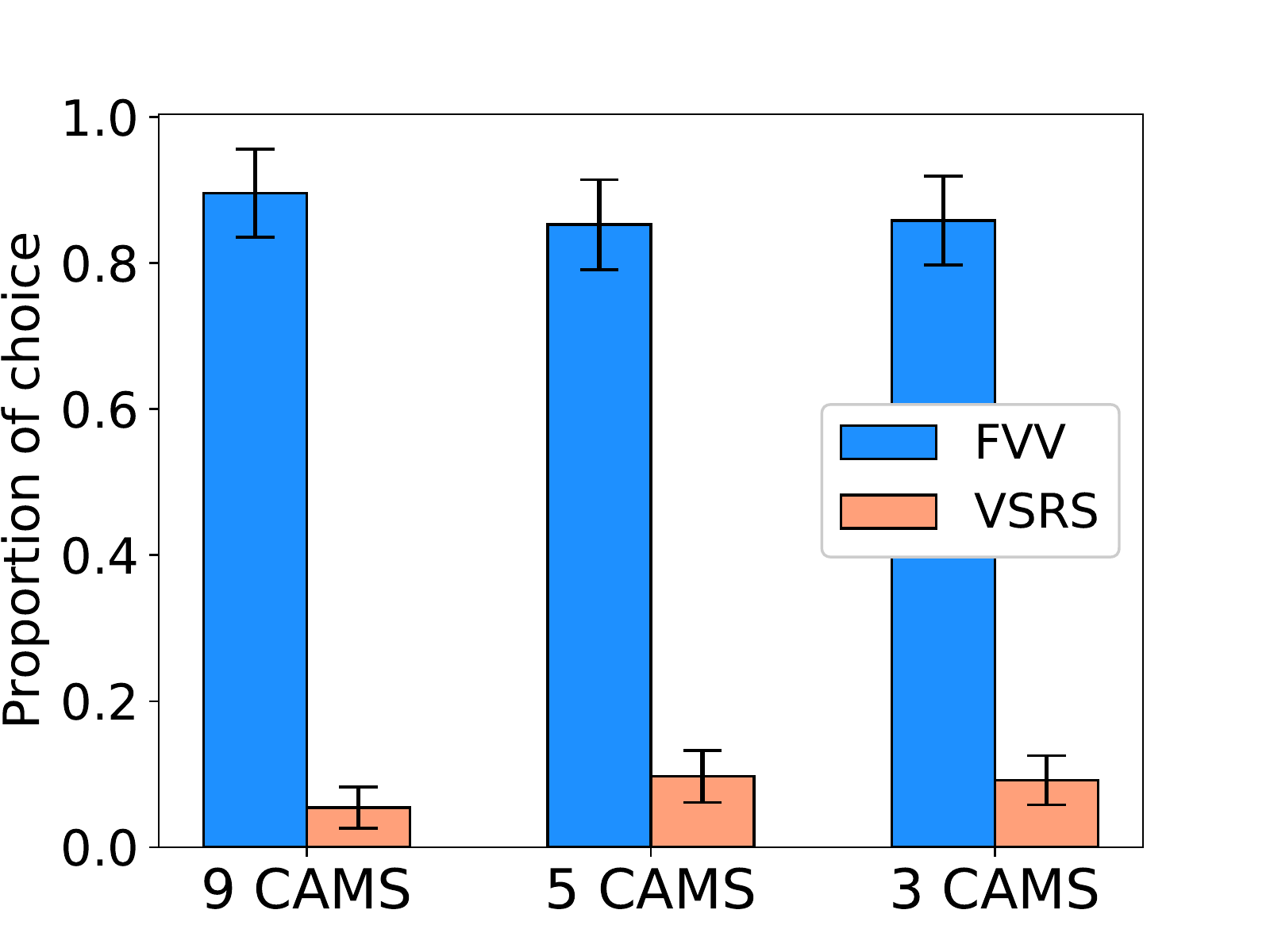}}
\subfloat[Trajectories\label{fig:PC-trajectory}]{\includegraphics[width=0.49\columnwidth]{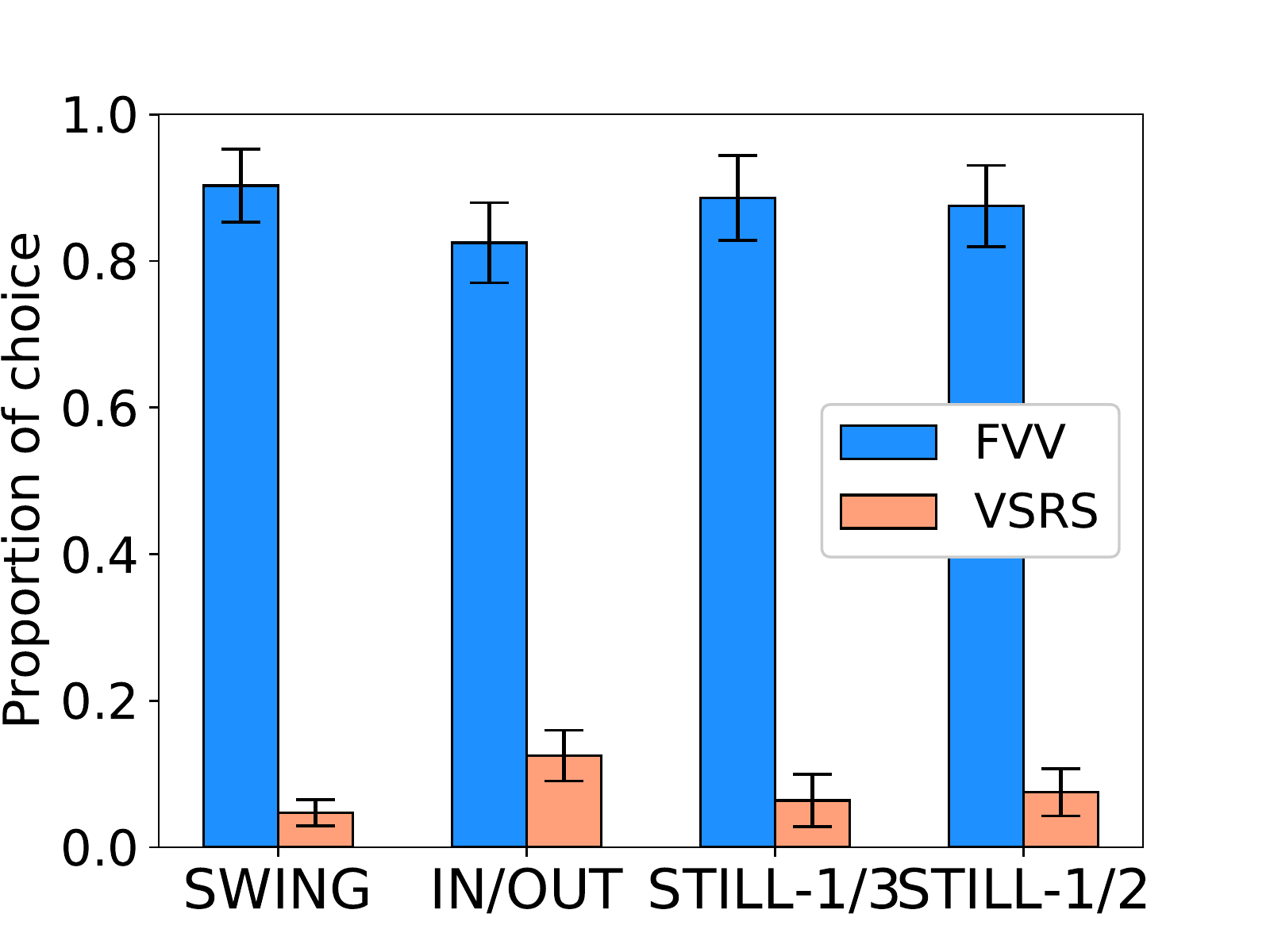}} 
\subfloat[Content\label{fig:PC-scenario}]{\includegraphics[width=0.49\columnwidth]{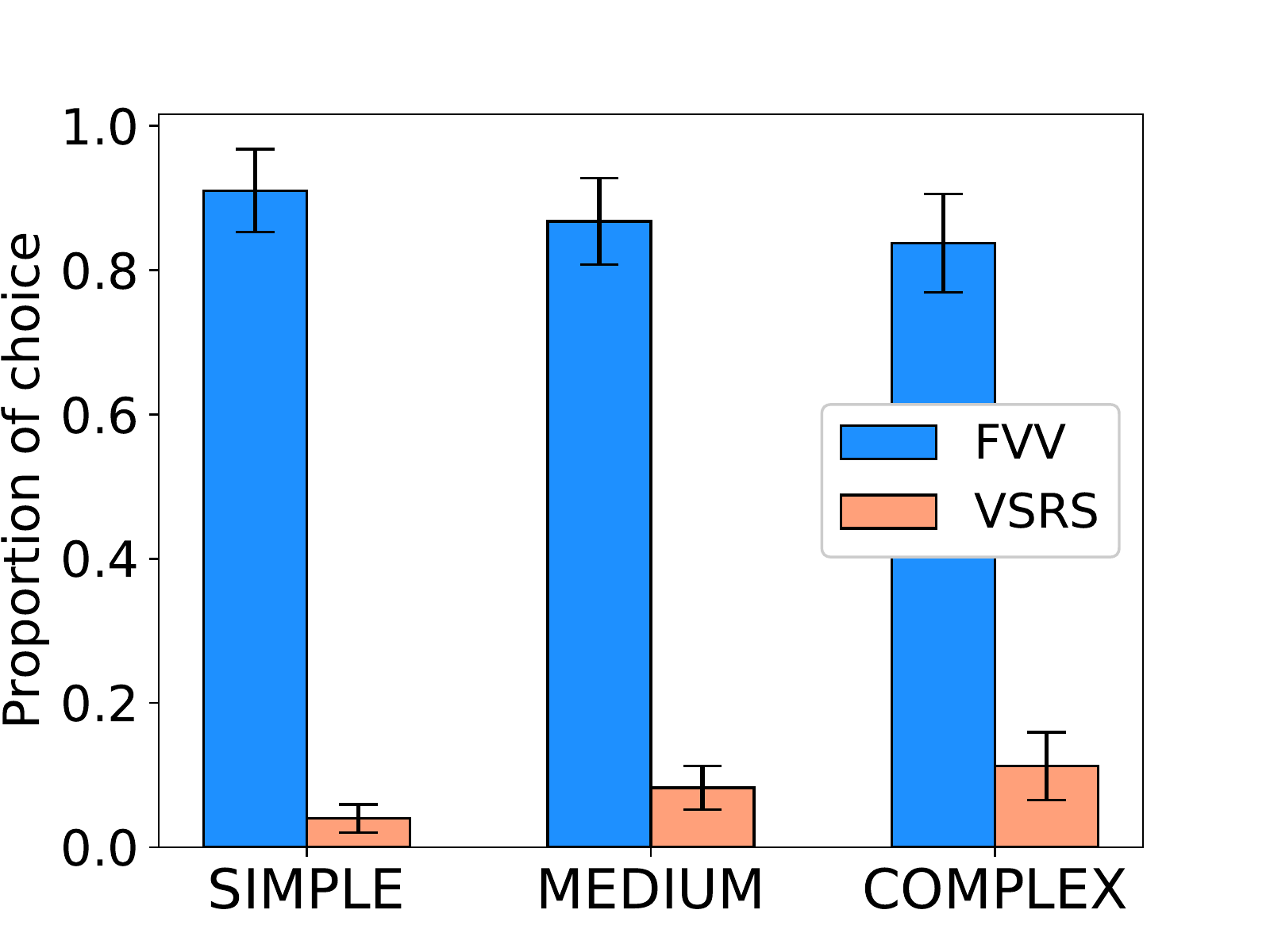}}
\caption{\label{fig:PC-results} Results of the comparative quality analysis (Pair Comparison): preference ratios for \SystemName\ (blue) and VSRS (orange)  }
\end{figure*}

\begin{figure*}[t]
\centering
\subfloat[Average MOS (all cases)\label{fig:ACR_General_MOS}]{\includegraphics[width=0.49\columnwidth]{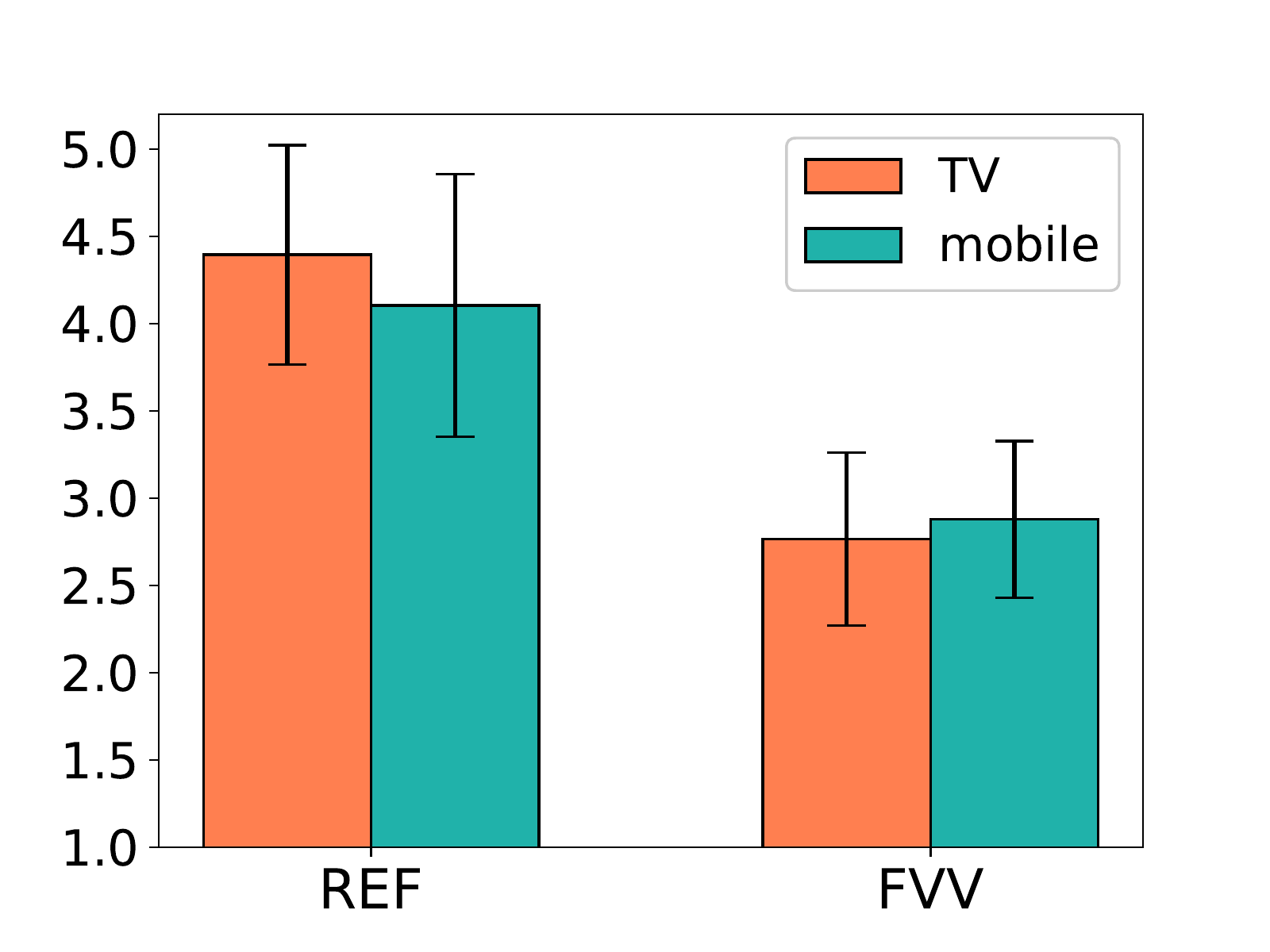}}
\subfloat[Average DMOS (all cases)\label{fig:ACR_General_DMOS}]{\includegraphics[width=0.49\columnwidth]{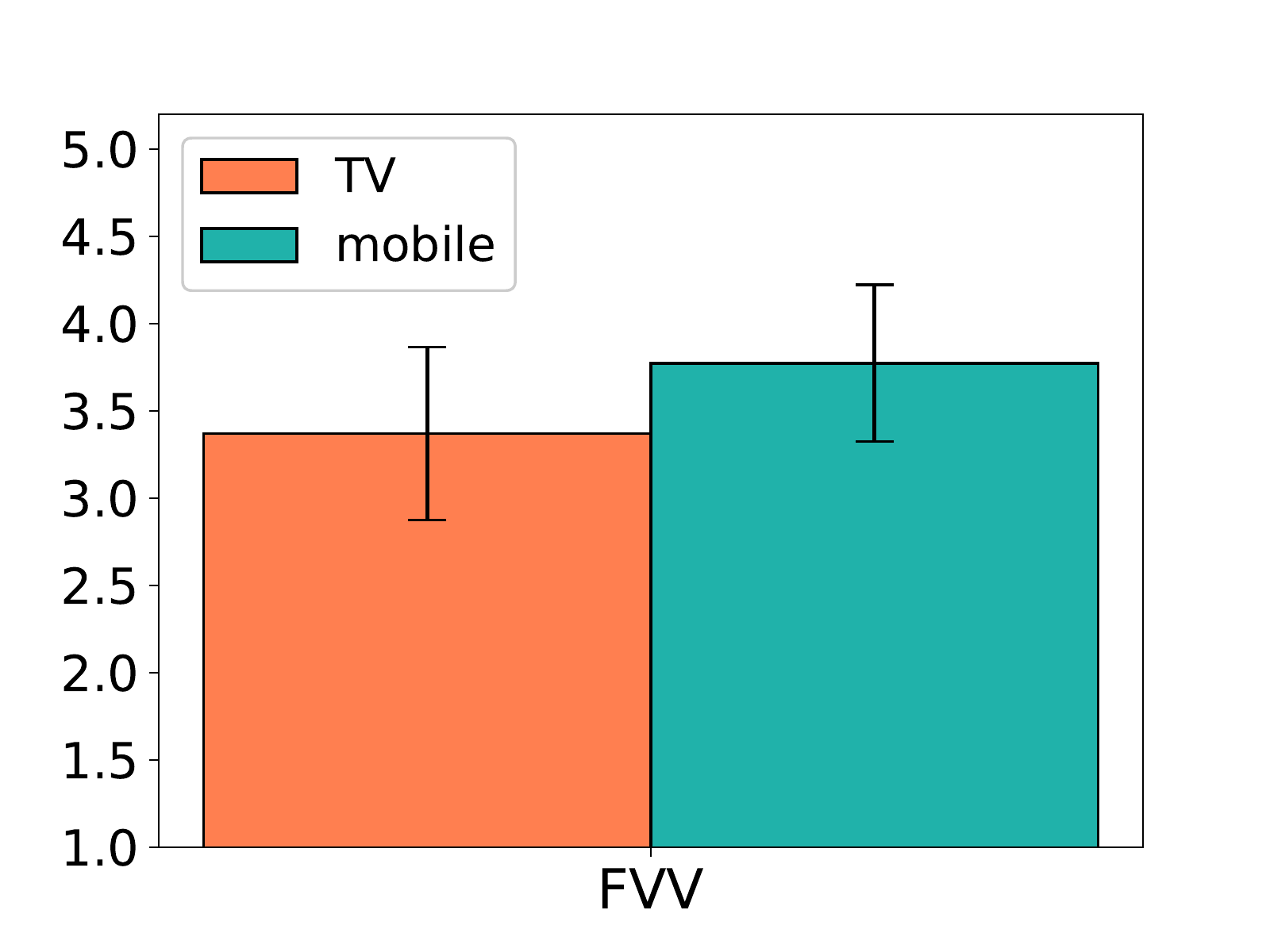}}
\subfloat[DMOS: Content\label{fig:DMOS_per_scenario}]{\includegraphics[width=0.49\columnwidth]{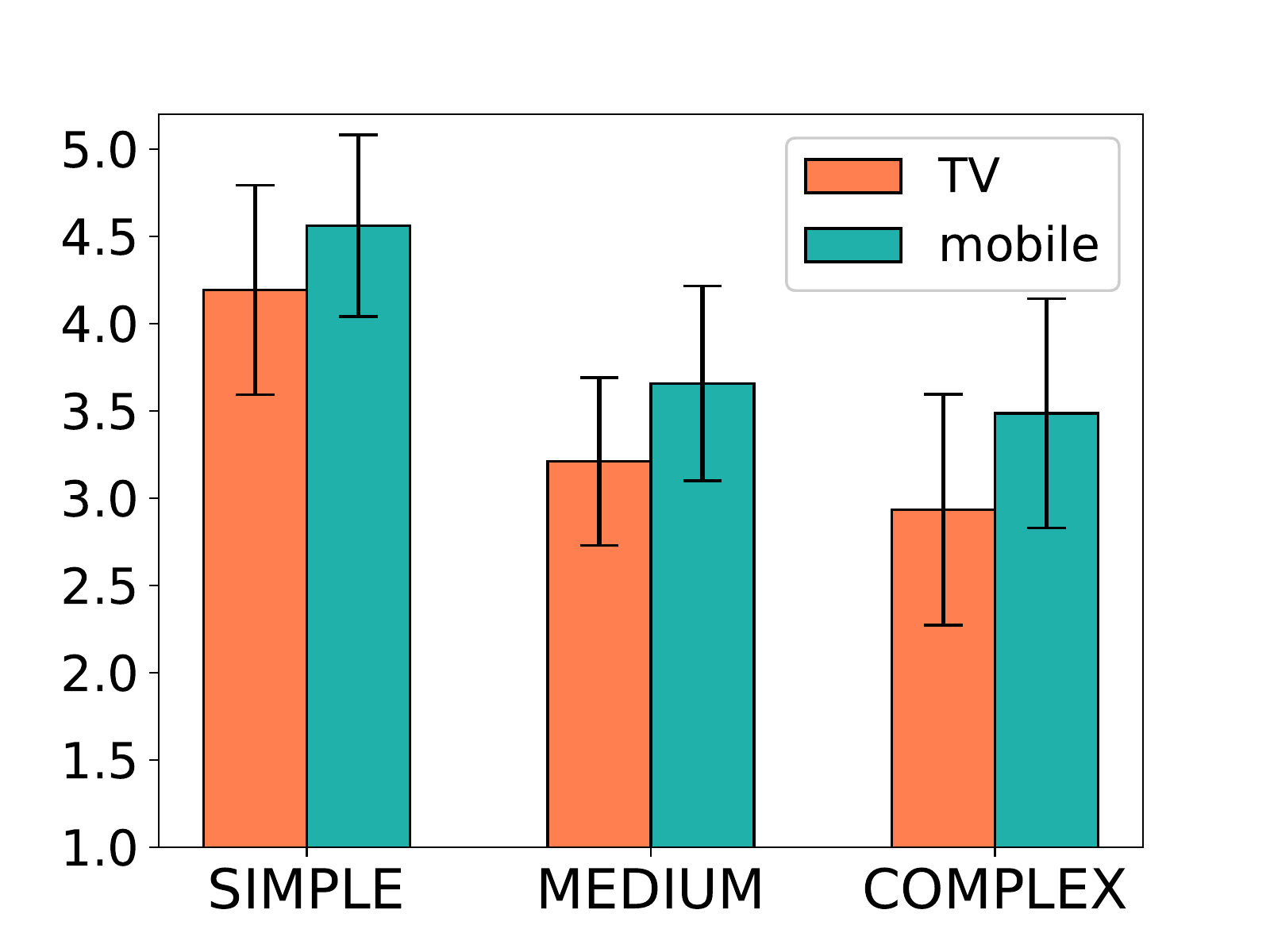}}
\subfloat[DMOS: Trajectories \label{fig:DMOS_per_virtual_camera_position}]{\includegraphics[width=0.49\columnwidth]{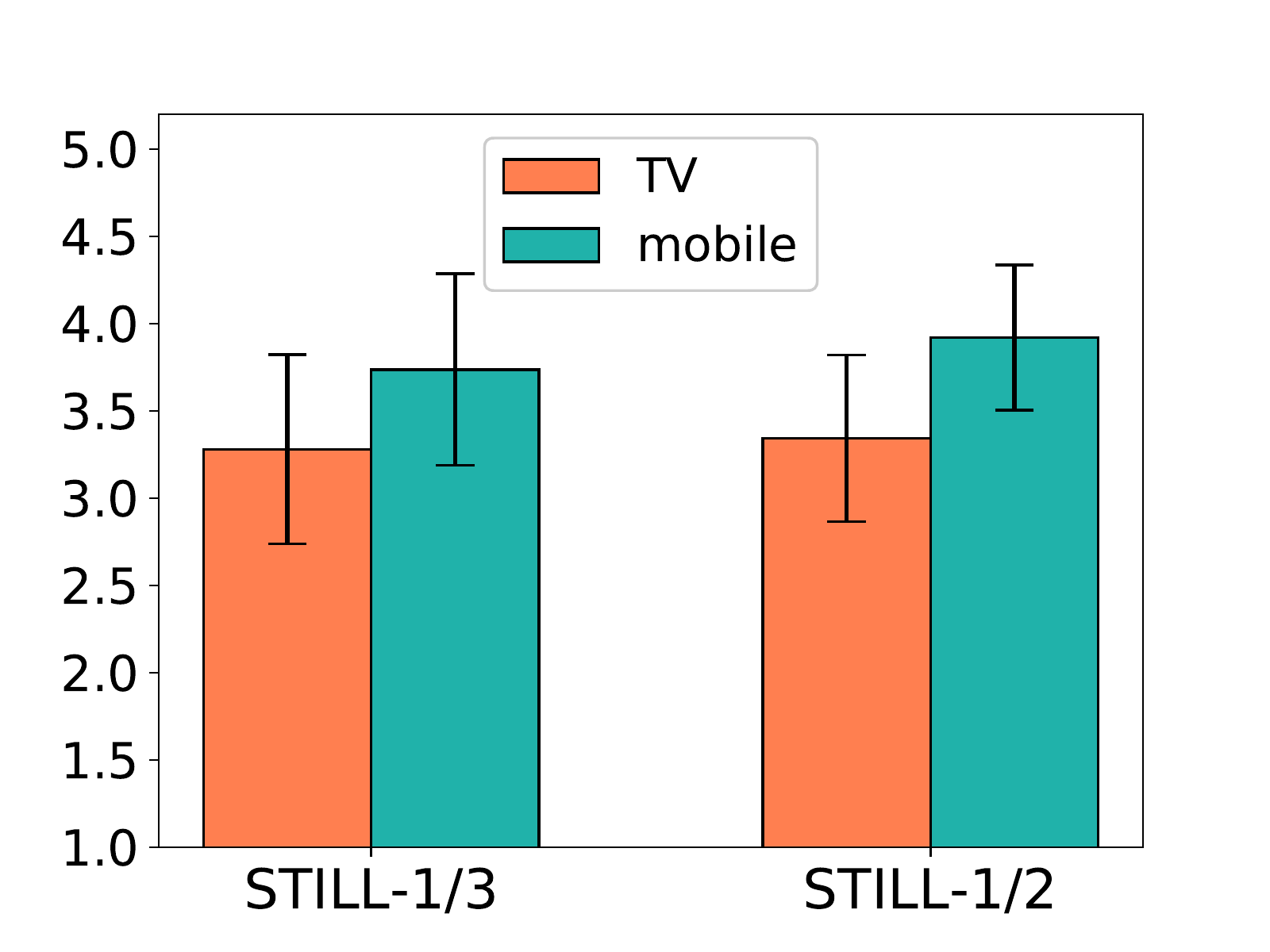}}
\caption{\label{fig:ACR-HR-DMOS-results} MOS and DMOS results of the absolute quality analysis (ACR-HR) for the PVS generated with the \SystemName\ synthesis. (a) Average MOS results are provided for the PVS and REF clips. (b)-(d) DMOS values calculated with respect to the average MOS of REF clips}
\end{figure*}

\subsubsection{Virtual camera trajectory} Still and moving virtual camera trajectories are generated (see Fig.~\ref{fig:camera_setting}) to evaluate the influence in the perceived quality of: i) position of the virtual view with respect to the reference cameras, and ii) camera movement.
\begin{itemize}
\item \textbf{Swing trajectory}: the virtual camera travels from the first to the last reference camera and back. 
\item \textbf{Step in-out trajectory}: the virtual camera steps in and out from the scene, starting from the reference camera in the middle of the arc.
\item \textbf{Still camera 1/2}: the virtual viewpoint is fixed half way between two reference cameras.%
\item \textbf{Still camera 1/3}: the virtual viewpoint is fixed between two reference cameras at 1/3 baseline distance from one of them.
\end{itemize}

\subsection{Equipment, environment and participants}
\noindent Two devices were used to perform the tests: 
\begin{itemize}
    \item TV set with a 55-inch screen and 4K (2160p) resolution (Samsung UE55HU8500)
    \item Smartphone with a 5.9-inch screen and FullHD (1080p) pixel resolution (Huawei Mate 9) 
\end{itemize}

The TV set was used in both tests, whereas the smartphone was only employed in the absolute quality analysis. For the TV set, in accordance to Rec. ITU-R BT. 500-13~\cite{methodologySubjective}, the viewing distance (VD) was set to 3H (3 times the screen height). For the smartphone, following the guidelines of Rec. ITU-T P.913~\cite{methodologySubjectiveP}, observers could select a comfortable VD and whether to hold the device or leave it on a table. 

The test area was set according to Rec. ITU-R BT. 500-13~\cite{methodologySubjective}, including the ambient lighting conditions. Specifically, brightness was set to around 25~lx to avoid disturbing reflections.

A total of 19 observers participated in both tests (5 women, 14 men), all of them having normal or corrected vision (glasses or lenses). The ages of the participants were between 23 and 36 (26.5 years on average).

\subsection{Comparative quality analysis: methodology}
\noindent The pair comparison (PC) methodology~\cite{itu2008subjective} was used. Pairs of corresponding PVS, generated using the \SystemName\ and VSRS synthesis methods respectively, are presented side by side (synchronously) in the TV display, covering the whole display’s width. This method allows the viewer to comparatively evaluate the synthesis impairments of both methods. The observers are asked to select the one they prefer. 

Following Rec. ITU-T P.910~\cite{itu2008subjective}, side-by-side sequences were padded
with mid-gray level to fit the screen size vertically. The 3H VD allows viewers to have a simultaneous view of both complete images, without requiring head movements.

Clips for all combinations of camera configuration, virtual camera trajectory and content were generated (only the 9-cam configuration was used for the Still Camera 1/3), resulting in 30 PVS pairs. Following Rec. ITU-T P.910~\cite{itu2008subjective}, all PVS pairs were presented in random order, and twice, alternating the left-right position in both repetitions. Including voting times, the duration of the test was approximately 16 minutes.

\subsection{Absolute quality analysis: methodology}
\noindent The absolute category rating with hidden reference (ACR-HR) method~\cite{itu2008subjective} with a 5-level scale (’bad’, ’poor’, ’fair’, ’good’, and ’excellent’) was used to quantify the visual quality of virtual views in the \SystemName\ system. Video sequences captured by physical cameras were used as hidden references (REF). They provide an upper bound for quality scores, as reference viewpoints do not present synthesis artifacts, but their visual quality is only limited by impairments caused by limitations of the cameras and video compression.  

To allow a fair comparison between PVS and REF (fixed viewpoints), only still camera trajectories were used. All contents in the dense 9-cam configuration were evaluated. PVSs are generated between cameras 3-4 and 4-5 at the positions defined by the Still Cameras 1/2 and 1/3 trajectories (see Fig.~\ref{fig:camera_setting}). Cameras 3 and 4 are used as REF. 

In the TV display, video sequences were presented at their original resolution, padded with mid-gray level to fit the screen size. Regarding the smartphone, video sequences cover the whole display. All clips were presented once in random order. Including voting times, the duration of the test was approximately 7.5 minutes.

\subsection{Comparative quality analysis: results and discussion}

\noindent Fig.~\ref{fig:PC-results} shows the results of the comparative quality analysis. Specifically, the preference ratios of both synthesis methods (\SystemName\ and VSRS). Mean values and 95\% Confidence Intervals are provided. The results show that the \SystemName\ synthesis clearly outperforms VSRS for all scenarios, trajectories and reference camera configurations. In average, the \SystemName\ synthesis is preferred over VSRS in well above 80\% of the comparisons, regardless of the trajectory or scenario. Preference values around 90\% is only very slightly decreased with an increasing complexity of the scenario (see Fig.~\ref{fig:PC-ncameras}) and a decreasing density in the camera setting (see Fig.~\ref{fig:PC-scenario}). Additionally, the preference for \SystemName\ is slightly lower for the Step in/out trajectory than the other cases (Swing and Still Camera).

\subsection{Absolute quality analysis: results and discussion}

\noindent Fig.~\ref{fig:ACR-HR-DMOS-results} shows the Mean Opinion Score (MOS) and Difference Mean Opinion Score (DMOS) of the absolute quality analysis, together with their respective 95\% Confidence Intervals. Average MOS results in Fig.~\ref{fig:ACR_General_MOS} are provided for the PVS and REF clips. MOS values for REF clips indicate that the quality of the physical cameras is limited (values below 4.5 for the TV and mobile). The DMOS values, computed using MOS values of REF clips as reference, are over 3 (fair quality) in all cases, regardless of device, trajectory or scenario. Maximum DMOS values sit in the range of [4-4.5] (close to indistinguishable from the quality of REF clips) for the Simple Scenario case, whereas a similar quality value in the range of [3-3.5] (between fair and good) is perceived for the Medium and Complex cases. It can be concluded that the user perception improves in all the cases when the content is presented onto a smaller screen (smartphone) where the synthesis artifacts are less visible. Finally, there is no statistical difference between the quality at different virtual camera postions (Still Camera 1/2 and 1/3), concluding that a stable visual quality is perceived along the virtual camera path.

\section{Conclusions}
\noindent We have presented \SystemName, a novel end-to-end free-viewpoint video system, designed for low deployment cost and real-time operation, only using off-the-shelf components. 

The paper describes the system architecture, including its functional blocks and details on the implementation of a complete system with nine Stereolabs ZED cameras, and consumer-grade servers and GPUs. The functional blocks of the system include tools to increase the quality of virtual views rendered from multiview plus depth data yielded by consumer-grade cameras. This includes depth correction in the capture servers, a lossless coding scheme for the transmission of high-precision depth data using off-the-shelf 8-bit video codecs, and layered view synthesis.

The visual quality of \SystemName\ has been assessed by means of a subjective evaluation with different conditions in free-viewpoint trajectories, camera configurations and content complexity. The results show that \SystemName\ outperforms VSRS for all conditions, and that the visual quality of the overall system is satisfactory, given the upper bound of the visual quality provided by the physical ZED cameras. With a mean motion-to-photon delay below 50~ms and an end-to-end delay of approximately 250~ms, \SystemName\ allows for seamless free navigation and bilateral immersive communications. 

\bibliographystyle{IEEEtran}
\bibliography{Bibliography_abbrev}

\end{document}